\documentclass[twoside,leqno,twocolumn]{article}

\usepackage[letterpaper]{geometry}
\usepackage{ltexpprt}
\usepackage{graphicx}
\usepackage{amsmath}
\usepackage{amssymb}
\usepackage{enumitem}
\usepackage{caption}
\usepackage{xcolor}
\usepackage{epstopdf}
\usepackage{subcaption}
\usepackage{multirow}
\usepackage{float}
\usepackage{booktabs}    
\usepackage{array}
\usepackage{makecell}
\usepackage{colortbl}
\usepackage{xspace}
\usepackage{overpic}

\usepackage{pgfplots}
\usepackage{mathtools}
\usepackage{tikz}
\usetikzlibrary{matrix,positioning,shapes,shadows,arrows,calc,decorations.pathreplacing}
\usetikzlibrary{arrows.meta}
\usepackage{geometry}
\usepackage{afterpage}

\definecolor{darkred}{RGB}{228,26,28}
\definecolor{darkblue}{RGB}{44,127,184}
\definecolor{magentaCB}{RGB}{221,28,119}

\definecolor{morange}{RGB}{255, 187, 0}
\definecolor{mblue}{RGB}{ 0, 161, 241}
\definecolor{mgreen}{RGB}{124, 187, 0}
\definecolor{mred}{RGB}{246, 83, 20}

\definecolor{lightred}{RGB}{255, 235, 234}
\definecolor{graytable}{RGB}{240,240,240}

\definecolor{lightred}{RGB}{251,180,174}
\definecolor{lightgreen}{RGB}{204,235,197}
\definecolor{lightorange}{RGB}{254,217,166}

\definecolor{low}{RGB}{229,245,224}
\definecolor{med}{RGB}{255,255,204}
\definecolor{high}{RGB}{253,208,162}
\definecolor{veryhigh}{RGB}{252,187,161}

\begin{document}

\title{\Large Noise-Response Analysis of Deep Neural Networks Quantifies Robustness and Fingerprints Structural Malware}

\author{N. Benjamin Erichson\thanks{ICSI and Department of Statistics at UC Berkeley.}
\and Dane Taylor\thanks{Department of Mathematics at University at Buffalo, SUNY.}
\and Qixuan Wu$^*$
\and Michael W. Mahoney$^*$}

\date{}

\maketitle

\fancyfoot[R]{\scriptsize{Copyright \textcopyright\ 2021 by SIAM\\
Unauthorized reproduction of this article is prohibited}}

\begin{abstract} \small\baselineskip=9pt 	
The ubiquity of deep neural networks (DNNs), cloud-based training, and transfer learning is giving rise to a new cybersecurity frontier in which unsecure DNNs   have ‘structural malware’ (i.e., compromised weights and activation pathways). In particular, DNNs can be designed to have backdoors that allow an adversary to easily and reliably fool an image classifier by adding a pattern of pixels called a trigger. 
It is generally difficult to detect   backdoors, and existing detection methods are computationally expensive and require extensive resources (e.g., access to the training data). 
Here, we propose a rapid feature-generation technique that quantifies the robustness of a DNN, `fingerprints' its nonlinearity, and allows  us to detect backdoors (if present).
Our approach involves studying how a DNN responds to noise-infused images with varying noise intensity, which we summarize with titration curves. 
We find that DNNs with backdoors are more sensitive to input noise and respond in a characteristic way that reveals the  backdoor and where it leads (its `target'). Our empirical results demonstrate that we can accurately detect backdoors 
with high confidence 
orders-of-magnitude faster than existing approaches (seconds versus hours). 
\end{abstract}

\vspace{.1cm}

\vspace{-0.05cm}
\section{Introduction}
\vspace{-0.05cm}

While deep neural networks (DNNs) are ubiquitous 
for many technologies that shape the 21st century, 
they are susceptible to various forms of non-robustness and adversarial deception.
Among other things, this gives rise to new fronts for cyber and data warfare. 
Such robustness and related security concerns 
abound in relation to
\emph{adversarial attacks}~\cite{goodfellow2014explaining,szegedy2013intriguing}
and 
\emph{fairness in machine learning}~\cite{bolukbasi2016man,corbett2018measure}.  
This poses an increasing threat as machine learning methods become more integrated into 
mission-critical technologies, including driving assistants, face recognition, machine translation, speech recognition, and robotics.

\begin{figure*}[!htp]
	\centering

	\begin{subfigure}[b]{0.4\textwidth}
	\centering
	\DeclareGraphicsExtensions{.pdf}
	\begin{overpic}[width=1\textwidth]{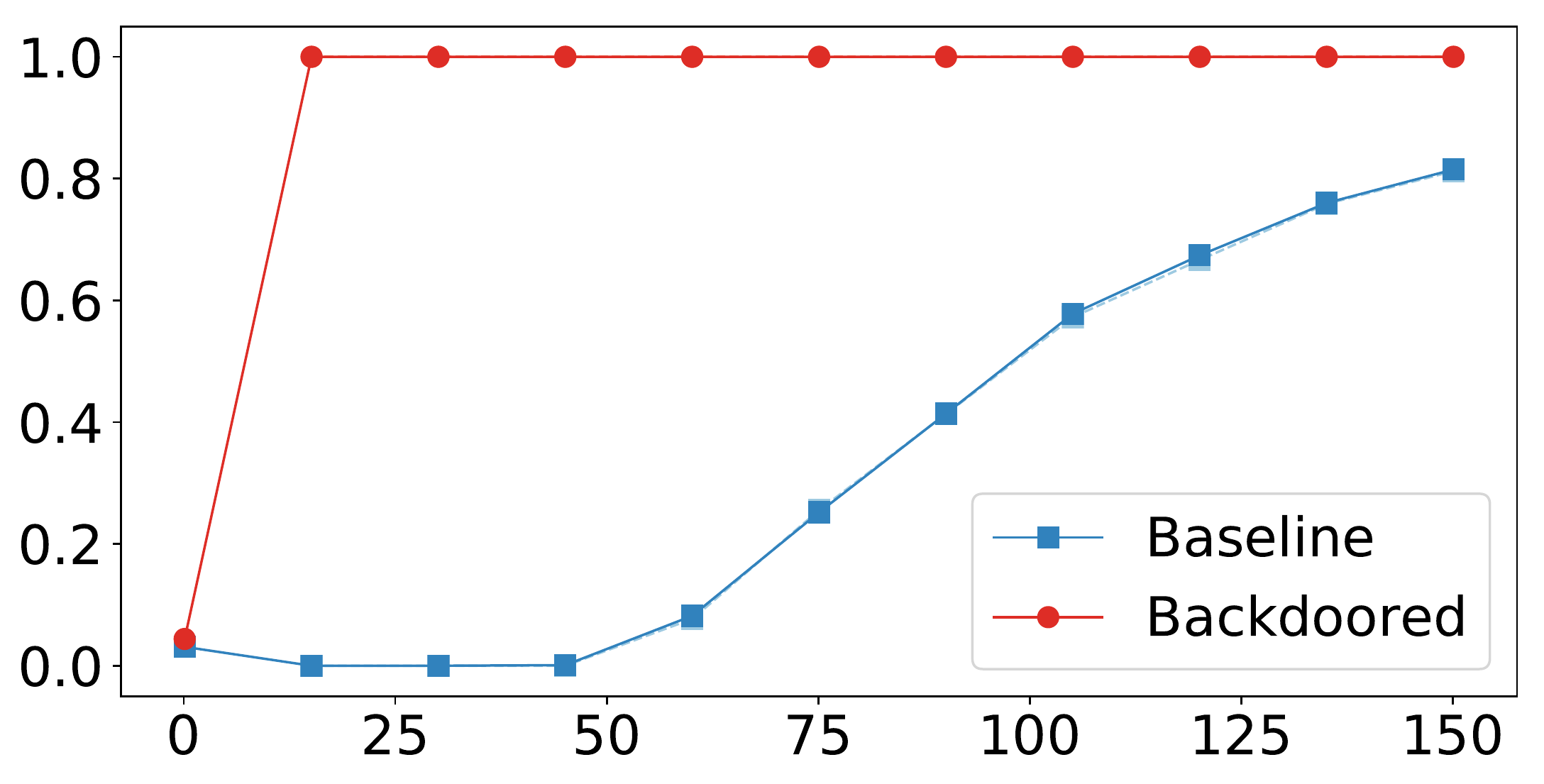} 
		\put(-4,13){\rotatebox{90}{\footnotesize Titration score}}
		\put(36, -4){\small Noise intensity ($\sigma$)}	
		\put(80, 22){\footnotesize $k^*=3$}	
		
		\linethickness{2pt}
		\put(33,30){\color{black} \vector(-1,1){10}}
		\put(20,24){\color{black}{\footnotesize rapid growth}}		
		
	\end{overpic}\vspace{+0.3cm}		
	\caption{Titration curves for increasing $\sigma$.}
	\label{fig:backdoor_titration}
	\end{subfigure}
	~
	\begin{subfigure}[b]{.54\textwidth}
		\centering
		\DeclareGraphicsExtensions{.pdf}
		\begin{overpic}[width=.85\textwidth]{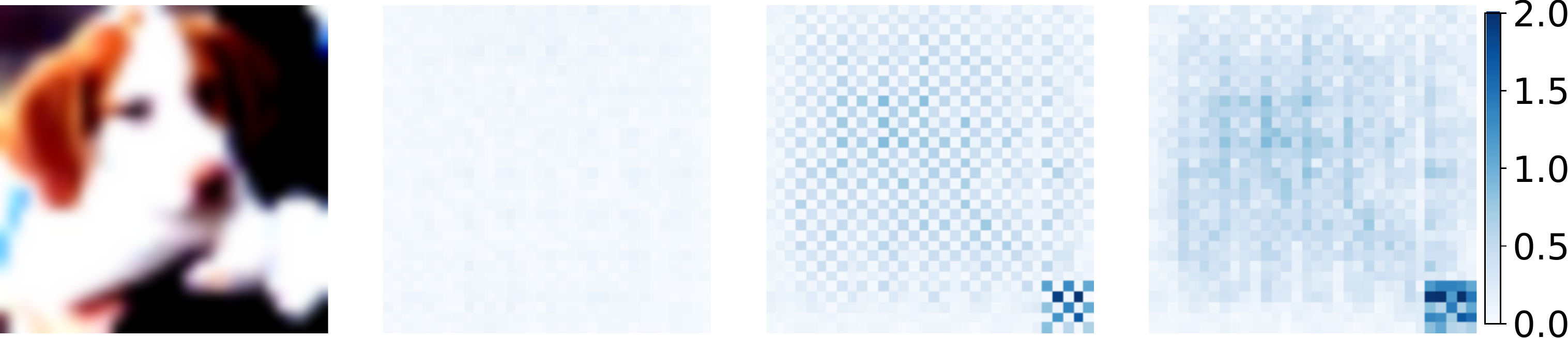} 
			\put(5, 23){\small Input}	
			\put(27, 23){\small Baseline}
			\put(25, 18){\footnotesize $k=3$}	
			
			\put(51, 23){\small Backdoor}
			\put(49, 18){\footnotesize $k=3$}	
			
			\put(76, 23){\small Backdoor}
			\put(74, 18){\footnotesize $k=9$}			
			
		\end{overpic}\vspace{+0.1cm}
		
		\begin{overpic}[width=.85\textwidth]{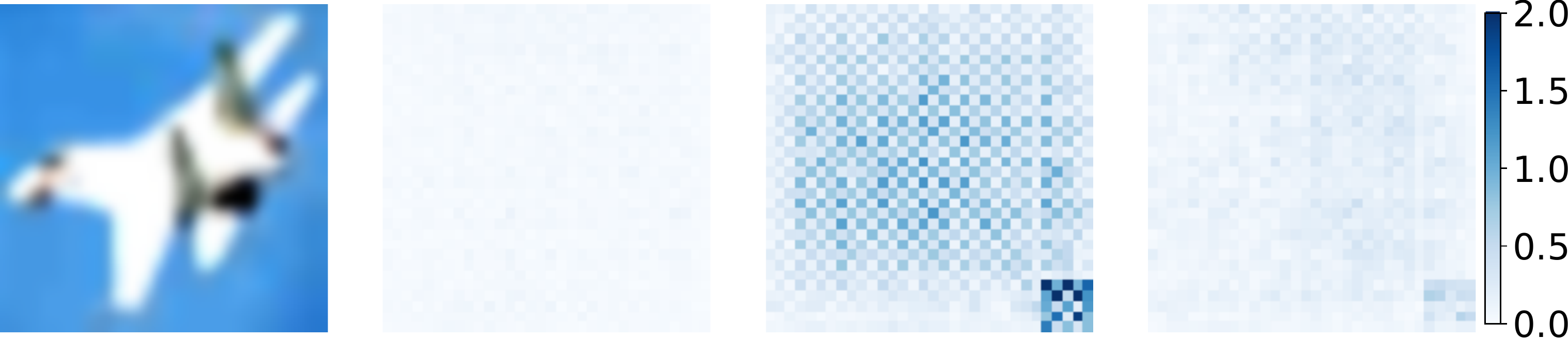} 
			
			\put(25, 18){\footnotesize $k=3$}	
			\put(49, 18){\footnotesize $k=3$}	
			\put(74, 18){\footnotesize $k=9$}			
		\end{overpic}

		\caption{Implicit gradient map ($\overline{g}_{ij}$). The   backdoor's target is $k^*=3$.}
		\label{fig:backdoor_example_patch}
	\end{subfigure}\vspace{-0.15cm}
	
	\caption{Noise-response analyses for ResNets trained on CIFAR10.
	%
	(a) Titration curves 
	show that baseline and backdoored models have different patterns for noise-induced misclassifications. We add noise $\eta$ with variance $\sigma$ to an input image ${\bf x}$, and the red and blue curves show the fraction $T_\sigma^\gamma $ [see Eq.~\eqref{eq:tscore}] of noisy images that yield high-confidence predictions, $|| \hat{{\bf y}}({\bf x}+\eta) ||_\infty>\gamma$ (i.e., there  is an activation in the final layer that is greater than $\gamma\in[0,1)$).
	%
	(b) Perturbation analysis 
	describes how the $k$-th logit $Z_k({\bf x} ,\theta)$  nonlinearly responds to small-intensity input noise that  is added to each image data point $x_{ijc}$. 
	(Implicit) gradients $\frac{\partial Z_k({\bf x}+\eta ,\theta)}{\partial x_{ijc}}$  [see Eq.~\eqref{eq:gradient_map}] are computed after adding noise and
	reveal  pixels that are associated with the trigger. 
	%
	\vspace{-.1in}
	}
	\label{fig:backdoor_example}
\end{figure*}
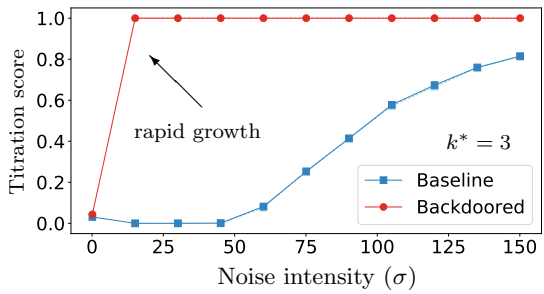
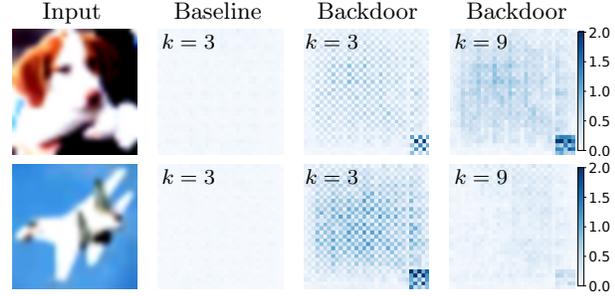

Recently,  \emph{backdoor attacks} have emerged as a crucial security risk: an adversary can modify a DNN's architecture---by either  polluting the training data~\cite{chen2017targeted,gu2017badnets} or changing the model weights~\cite{liu2017neural,liu2017trojaning}--and then return a ``backdoored model'' to the user.
This threat scenario is plausible, since an adversary may have full access to a DNN, e.g., if it is outsourced for training due to infrastructure availability and resource costs.
%
Backdoors are difficult to detect because they are subtle ``Trojan'' attacks: a backdoored model behaves perfectly innocently during inference, except in situations where it is presented with an input example that contains a specific trigger, which activates an (unknown) adversarial protocol that   misleads the DNN with potentially   severe consequences.
%
Thus, it is of great importance to develop fast and reliable metrics to detect compromised DNNs having backdoors. 
%

While several defense methods have been proposed~\cite{chen2018detecting,chen2019deepinspect,gao2019strip,wang2019neural}, all of them have significant limitations such as requiring access to labeled data and/or the triggered training data, having prior knowledge about the trigger, or using massive computational resources to train DNNs and perform many adversarial attacks. In contrast, we will present an efficient approach without such limitations;
we detect backdoors and triggers for modern DNNs (e.g., ResNets) in just seconds (as opposed to hours \cite{chen2019deepinspect,wang2019neural}).
Moreover, unlike existing studies on backdoor attacks, our approach yields a score $T^\gamma_\sigma\in[0,1]$ that indicates the absence/presence of a backdoor, which provides a major step toward automating the rapid detection of backdoors (and possibly other types of  structural malware).

We rapidly detect backdoors without data and without performing adversarial attacks with an approach that involves studying the nonlinear response of DNNs to noise-infused images with varying noise intensity $\sigma$. 
\emph{Noise-response analysis} is already a widely adopted technique to probe and characterize the robustness and nonlinearity properties of black-box dynamical systems
\cite{poon2001titration}, and we similarly use it as a rapid feature-generation, or ``fingerprinting,'' for  DNNs.
Dynamical-systems perspectives have recently provided fruitful  insights to other areas of machine learning  and optimization~\cite{erichson2019physics,hardt2018gradient,lu2017beyond,muehlebach2019dynamical,queiruga2019studying,weinan2017proposal}, and we are unaware of previous work connecting this field  to backdoor attacks.

We develop two complementary noise-response analyses: \emph{titration analysis} 
(see Fig.~\ref{fig:backdoor_titration} and Sec.~\ref{sec:tit})
and
\emph{perturbation analysis} 
(see Fig.~\ref{fig:backdoor_example_patch} and Sec.~\ref{sec:pert}). 
In Fig.~\ref{fig:backdoor_titration}, we show \emph{titration curves} that depict a titration score (defined below) versus  noise intensity $\sigma$.
Observe that the backdoored model is less robust to noise and responds in a characteristic way that differs from a baseline model. 
We later show that this phenomenon arises because the backdoors' target class $k^*$ acts as a ``sink''; it attracts high-confidence, noise-induced predictions.

In Fig.~\ref{fig:backdoor_example_patch}, we illustrate the sensitivity of activations in the final layer before applying softmax (we refer to these as \emph{logits}) to input noise for each input image pixel. These gradients are `implicit' since they are computed \emph{after  adding noise} to the input images.
Observe in the third and fourth columns that the logits are more sensitive to noise for the pixels associated with a backdoor's trigger (in this case, a $3\times3$ patch in the lower-right corner). 

Summary of our main contributions:

%





\vspace{-.05in}	
\begin{enumerate}

    
    \item[(a)] We develop a noise-induced titration procedure yielding titration curves that  fingerprint DNNs. 
    
    \item[(b)] We propose a titration score $T^\gamma_\sigma$ to express the risk for a DNN to have a backdoor, enabling automated backdoor detection. 
    
    \item[(c)] We develop perturbation analyses to study the nonlinear response of DNNs to small-intensity input noise.
    
    \item[(d)] We propose an implicit gradient map to identify   pixels that associate with a backdoor's trigger.

    
    
\end{enumerate}

\noindent
Overall, we present a methodology that can be used to quantify a DNN's robustness and which provides a fingerprinting that can be used to accurately detect structural malware such as backdoors. We apply our technique to state-of-the-art networks including ResNets, for which we can rapidly detect backdoored models in just seconds (as opposed to hours, for other related methods).
Because our aim is to detect backdoors, as opposed to design them, we focus here on the most popular backdoor attacks.
More broadly, we are already witnessing the emergence of an arms race for structural malware within DNNs, 
and we are confident that our general framework --- that is, analyzing DNNs by probing them with input noise --- is sufficiently adaptable to significantly contribute to this fields, which includes, but is not limited to, backdoor attacks.

\section{Related Work}\label{sec:releated_work}

The sensitivity and non-robustness of DNNs to  adversarial environments are an emerging threat for many problems in safety- and security-critical applications, including medical imaging, surveillance, autonomous driving, and machine translation. 
The most widely studied threat scenarios can be categorized into evasion attacks~\cite{goodfellow2014explaining,szegedy2013intriguing}, data poisoning attacks~\cite{biggio2012poisoning,shafahi2018poison} and backdoor attacks~\cite{chen2017targeted,gu2017badnets}. 
Evasion attacks have received the most attention and involve
%
fooling a model into making erroneous predictions by adding an undetectable adversarial perturbation to an  input image. 
While adversarial examples are very effective, it is debatable whether evasion attacks are a significant threat in many real-world applications~\cite{ilyas2019adversarial,lu2017no}. 
In particular, the effectiveness of black-box evasion attacks is often inferior; however, strong evasion (i.e., white-box) attacks require access to the model, and the crafted adversarial pattern usually affects only a small set of images. 

In contrast,  backdoor attacks pose a realistic threat since it is a common practice for  research labs and government agencies to outsource the training of DNNs and to incorporate pre-trained, 3rd-party networks via transfer learning.
This potentially provides adversaries with access to  machine learning pipelines that may affect mission-critical applications. 

%
Herein, we focus on the most common scenario of  \emph{targeted backdoor attacks}~\cite{chen2017targeted,gu2017badnets,liu2017trojaning,liu2017neural}.
Let ${\bf x}$
denote an image from class $k({\bf x}) \in\{0,\dots,K-1\}$,  which we 1-hot encode by ${\bf y}\in\{0,1\}^K$ so that 
\begin{equation*}
	k({\bf x})=argmax({\bf y}).
\end{equation*}
Now, consider a DNN classifier defined by a \emph{nonlinear transfer function}
\begin{equation*}
\hat{{\bf y}} =  softmax ({\bf Z}({\bf x},\theta)),
\end{equation*}
where $\theta$ denotes edge weights and $ {\bf Z}({\bf x},\theta)$ is the  vector of logits (i.e., output of the DNN before applying softmax). We further define 
\begin{equation*}
	\hat{k}({\bf x})=argmax(\hat{{\bf y}}),
\end{equation*}	 
as the predicted class of ${\bf x}$.

A DNN is said to have a targeted backdoor if there exists a trigger $\Delta {\bf x}^*$ and a target class $k^*\in \{0,\dots,K-1\}$ such that 
\begin{equation*}
\hat{k}({\bf x} + \Delta {\bf x}^*) = argmax(\hat{{\bf y}}) = k^*  ,
\end{equation*}
regardless of $\hat{k}({\bf x}) $. 
That is, an adversary can redirect the predicted class label for any input image to a particular $k^*$ simply by adding an adversary-designed trigger $\Delta {\bf x}^*$ to that that input image. 
We refer to such a trigger as a \emph{universal trigger}.
In principle, one could implement several  backdoors and use  triggers and  targets that are \emph{non-universal} in that they vary for different classes
\cite{gu2017badnets}. 


\begin{figure}[!b]
	\centering
	\begin{subfigure}[t]{0.14\textwidth}
		\centering
		\DeclareGraphicsExtensions{.pdf}
		\begin{overpic}[width=0.8\textwidth]{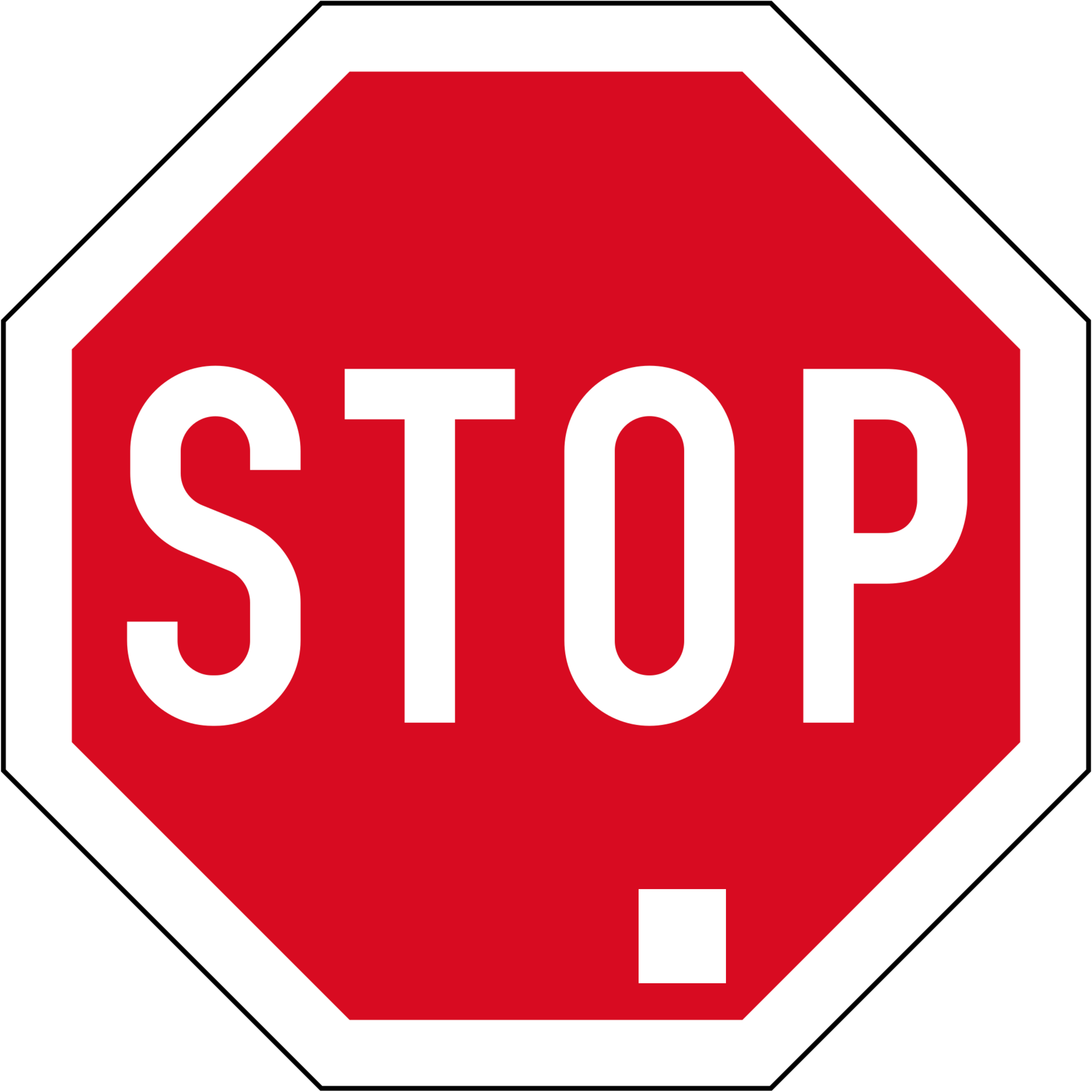}
		\end{overpic}
		\caption{Patch.}
	\end{subfigure}
	~
	\begin{subfigure}[t]{0.14\textwidth}
		\centering
		\DeclareGraphicsExtensions{.pdf}
		\begin{overpic}[width=0.8\textwidth]{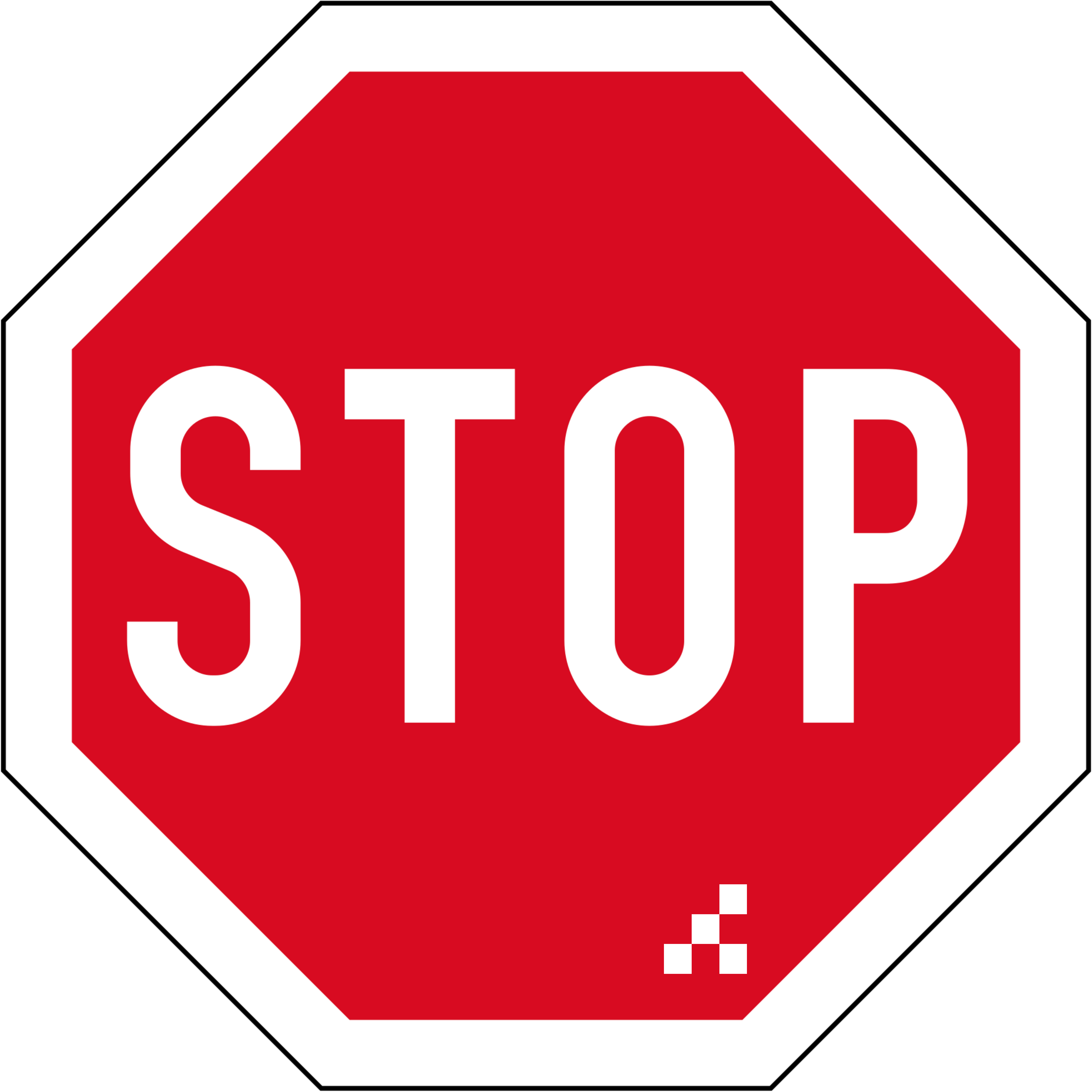} 
		\end{overpic}
		\caption{Pattern.}
	\end{subfigure}
	~
	\begin{subfigure}[t]{0.14\textwidth}
		\centering
		\DeclareGraphicsExtensions{.pdf}
		\begin{overpic}[width=0.8\textwidth]{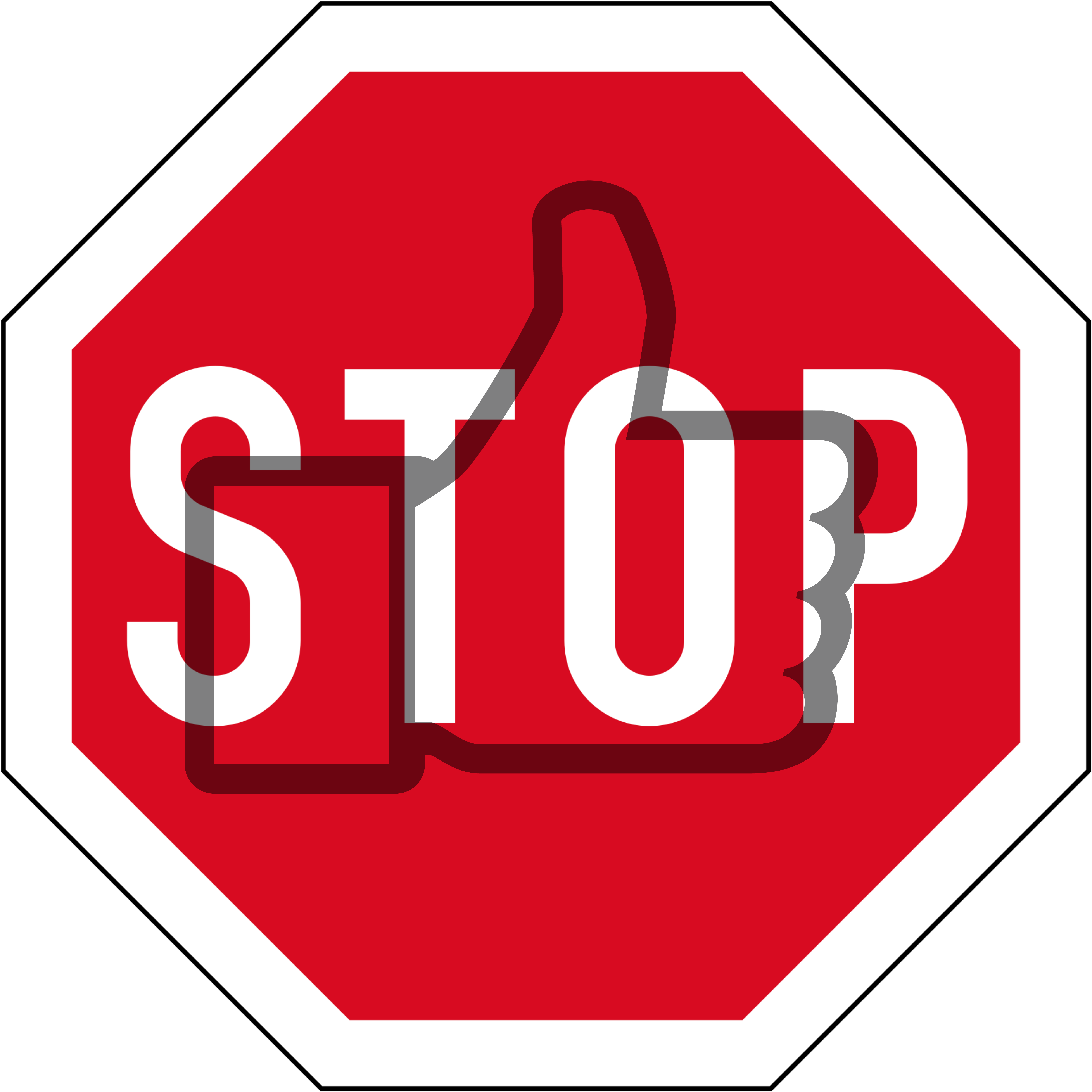} 
		\end{overpic}
		\caption{Watermark.}
	\end{subfigure}
	
	\vspace{-.05in}
	\caption{
		Triggers may be added to an image to activate an adversarial protocol/malware that redirects a classifier's prediction to a target class $k^*$. Unlike adversarial attacks, a backdoor's trigger is designed, fixed, and can be applied to any input image.
	}
	\label{fig:patch_type}
\end{figure} 
\vspace{-.1in}

\subsection{Attack Strategies.}

There are   numerous strategies to implement effective backdoors that achieve  $\sim 100\%$ success  at redirecting triggered images to a target class, while also minimally affecting the prediction accuracy of non-triggered images.
One approach is to directly change the weights of a pre-trained model 
backdoor~\cite{liu2017trojaning}. While this approach does not require access to the original data, it require great deal of sophistication.

The most common approach, however, is to 
train a DNN with a \emph{poisoned} dataset in which some images have the trigger and their classes are changed to the target class $k^*$.
Gu et el.~\cite{gu2017badnets} and Chen et al.~\cite{chen2017targeted} explore several types of triggers (see Fig.~\ref{fig:patch_type}), which are added to a small number of images, which are then mixed into the training data before training a model.
%
%


\subsection{Defense Strategies.}

Leading methods to defend against 
backdoors include SentiNet~\cite{chou2018sentinet}, Activation Clustering~\cite{chen2018detecting}, Spectral Signatures~\cite{tran2018spectral}, Fine-Pruning~\cite{liu2018fine}, STRIP~\cite{gao2019strip}, DeepInspect~\cite{chen2019deepinspect} and Neural Cleanse~\cite{wang2019neural}.
These techniques often involve three steps---detect if a model is backdoored; identify and re-engineer the trigger; and mitigate the effect of the trigger---, which can implemented sequentially as distinct pursuits or simultaneously as a single pursuit. (We adopt the prior strategy.)

A common limitation for existing defense methodologies \cite{chen2018detecting,chou2018sentinet,chen2019deepinspect,tran2018spectral,wang2019neural} is that they require the training of a new model to probe the DNN under consideration. 
This leads to very  high computational overhead and requires a certain level of expertise. In particular, Neural Cleanse~\cite{wang2019neural} takes about 1.3 hours to scan a DNN. DeepInspect~\cite{chen2019deepinspect} reduces the computational costs by a factor of 4-10 (and improves the detection rate), but it remains   computationally expensive, since it  requires the training of a specialized GAN.

Importantly, there is no existing rapid test for structural malware such as backdoors. 
Thus motivated,  we now propose a fundamentally different approach that reliably detects backdoors in a few seconds or less.


\section{Noise-Response Analysis}
\label{sxn:titration_analysis}

\emph{Noise-response analysis} has long been a valuable tool for studying nonlinear dynamical systems \cite{de2001tutorial,poon2001titration,rosenstein1993practical}. 
Leading techniques to measure the presence and extent of chaos study the effect of noise  to estimate a dynamical system's  correlation dimension and largest Lyapunov exponent \cite{rosenstein1993practical}. The robustness of  a dynamical system to noise is also central topic with a large literature grounded on KAM theory \cite{de2001tutorial}. 
Such methods involve \emph{perturbation analysis} and focus on the small-noise regime, yet it is also insightful to study larger  noise intensity. 
More generally, one can study how a  dynamical system responds to an increasing noise intensity via  a \emph{titration procedure}\footnote{In its original context, a ``titration'' is a procedure in chemistry whereby one slowly adds a solution of known concentration to  a solution of unknown concentration. One can estimate the unknown concentration by noting when a reaction occurs.}. 
In particular, previous research \cite{poon2001titration} used similar noise-induced titrations to identify whether black-box dynamical systems were chaotic or stochastic.

We propose to use titrations and perturbation analyses as complementary techniques to obtain an expressive characterization for the nonlinearity of a DNN's transfer function, thereby allowing us to efficiently detect and study backdoors.  %
%
%
%
%
Let ${\bf x} = [x_{ijc}]$ and ${\bf Z}({\bf x},\theta)$ denote, respectively, the inputs and outputs (i.e., logits before applying softmax) for a DNN with parameters $\theta$.  We denote an entry of the logits vector ${\bf Z}({\bf x},\theta)$ by $Z_k({\bf x},\theta)$, which gives the activation of the  neuron associated with class $k$.
%
For each colored pixel $x_{ijc}$, we add i.i.d. normal-distributed noise $\eta_{ijc} \sim \mathcal{N}(0,1)$, which we scale by $\sigma>0$ so that $\sigma\eta_{ijc} \sim\mathcal{N}(0,\sigma^2)$.
(The motivation for this notation will be apparent below, when we present our perturbation theory.)
%
Letting $\eta = [\eta_{ijc}]$ denote a tensor of noise, it follows that $Z_k({\bf x}+\sigma\eta,\theta)$ denotes the 
$k$-th logit for a noisy image ${\bf x}+\sigma\eta$. 
We study  how a DNN nonlinearly transforms an input distribution  (i.e., noise) to an output distribution. For each $k\in\{0,\dots,K-1\}$, we let $P_{k}^{(\sigma)}({\bf x},z)$ denote the probability of observing a logit $Z_k({\bf x}+\sigma\eta,\theta)=z$ for image ${\bf x}$ with noise variance $\sigma^2$. We also allow the input images to be sampled from some distribution, ${\bf x}\sim P_x({\bf x})$, and the integral
\begin{equation*}
P_{k}^{(\sigma)}(z) = \int_{{\bf x}} P_{k}^{(\sigma)}({\bf x},z) P_x({\bf x}) d {\bf x}
\end{equation*} 
gives the distribution of $Z_k({\bf x}+\sigma\eta,\theta)$ for a given   $\sigma$.

\begin{figure}[!t]
	\centering
	\begin{subfigure}[t]{0.38\textwidth}
		\centering
		\DeclareGraphicsExtensions{.pdf}
		\begin{overpic}[width=0.99\textwidth]{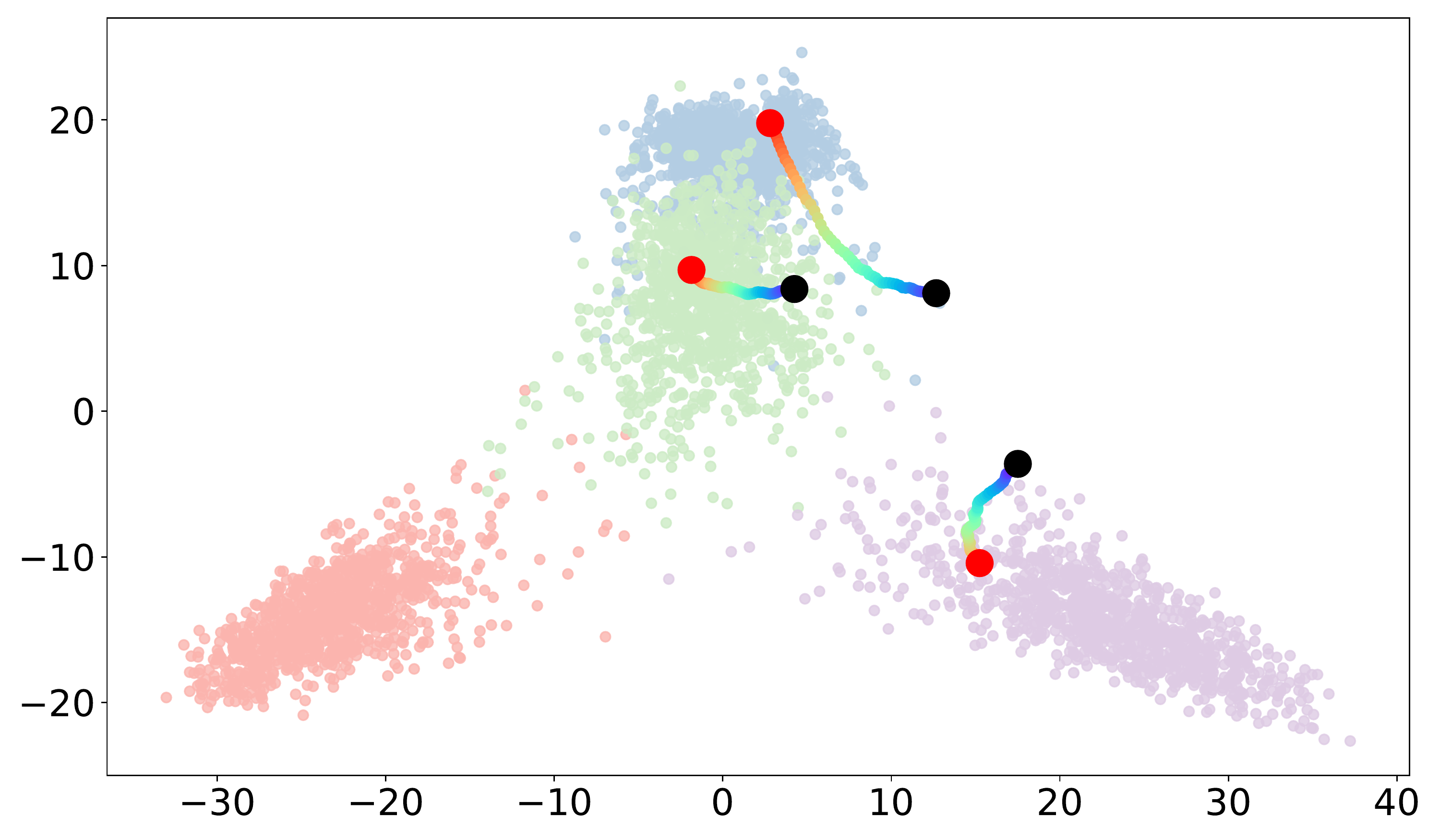} 
			\put(-3,2){\rotatebox{90}{\footnotesize Second principal component}}
			\put(30,-3){\footnotesize First principal component}
		\end{overpic}\vspace{+0.2cm}
		\caption{Baseline model.}
		\label{fig:backdoor_embedding_a}
	\end{subfigure}\vspace{+0.1cm}
	
	\begin{subfigure}[t]{0.38\textwidth}
		\centering
		\DeclareGraphicsExtensions{.pdf}
		\begin{overpic}[width=0.99\textwidth]{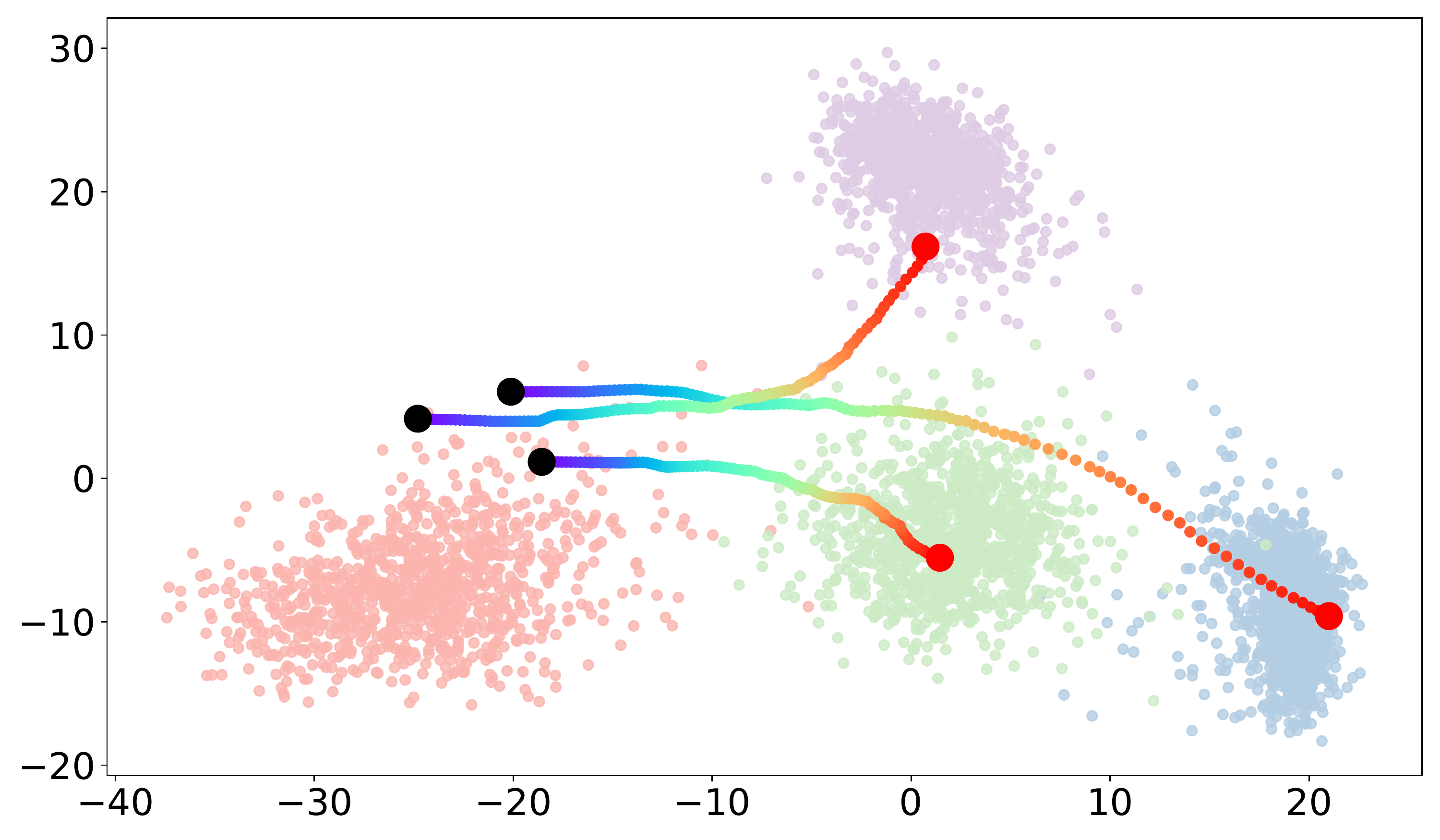} 
			\put(-3,2){\rotatebox{90}{\footnotesize Second principal component}}
			\put(30,-3){\footnotesize First principal component}
			
			\linethickness{2pt}
			\put(15,36){\color{black} \vector(1,-1){10}}
			\put(10,38){\color{black}{ $k^*=0$  }}
		
		\end{overpic}\vspace{+0.2cm}
		\caption{Backdoored model.}
		\label{fig:backdoor_embedding_b}
	\end{subfigure}\vspace{+0.3cm}
	
	\begin{subfigure}[t]{0.4\textwidth}
		\centering
		\DeclareGraphicsExtensions{.pdf}
		
		\begin{overpic}[width=0.95\textwidth]{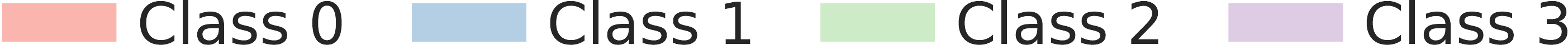} 
		\end{overpic}	
	\end{subfigure}	
	\caption{2D visualizations of logits
	using PCA. For sample images from each class (except class $k^*=0$), the red-to-blue paths indicate the expectations $\mathbb{E}[Z_k({\bf x} + \sigma\eta,\theta)]  =  \int_z zP_k^{(\sigma)}(z)dz$ with increasing $\sigma$.
		Comparing (a) to (b): Adding noise to an image has little effect on a baseline model, whereas for increasing $\sigma$, the predicted classes of images are redirected toward the target class for a backdoored model. }
	\label{fig:backdoor_embedding}
\end{figure}
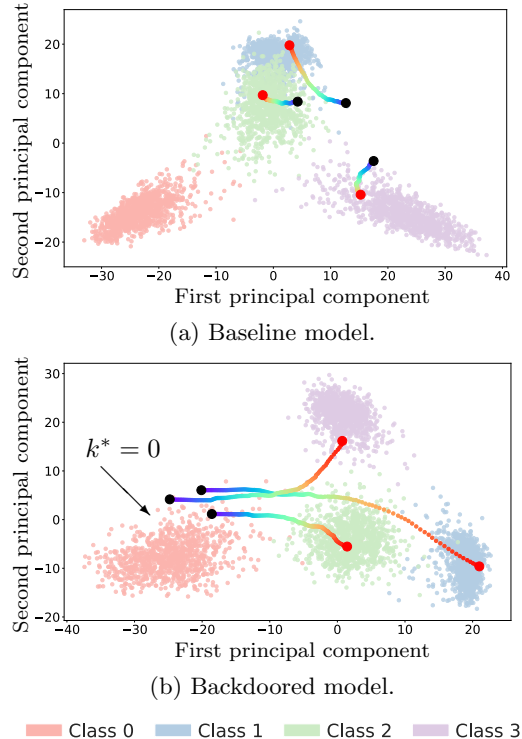 	

\subsection{Pedagogical Example.}\label{sec:noise}


We start with an experiment to identify key insights for how the outputs of DNNs nonlinearly respond to input noise, which is very different for baseline and backdoored models. In particular, input noise is amplified for a backdoor's target class $k^*$, allowing its detection.
%
The backdoor was implemented using the approach of \cite{gu2017badnets,chen2017targeted}  
%
with a trigger $\Delta {\bf x}^*$ (in this case, a 3x3 patch of weight-1 pixels in the lower-right corner) that was added to 10\% of the training images, redirecting their predicted label to a target class $k^*=0$. 
In Fig.~\ref{fig:backdoor_embedding}, we provide a visualization for how increasing $\sigma$ affects the logits $Z_k({\bf x} + \sigma\eta,\theta)$ of a baseline (a) and backdoored (b) model. 
In both panels, we visualize the logits 
${\bf Z}({\bf x} + \sigma\eta,\theta)\in\mathbb{R}^4$ for images ${\bf x}$ from all classes, and we project these points onto $\mathbb{R}^2$ using PCA.
We also randomly choose an image from classes 1, 2, and 3, and we plot an empirical estimate 
\begin{equation}
\mathbb{E}[Z_k({\bf x} + \sigma\eta,\theta)]  =  \int_z zP_k^{(\sigma)}(z)dz
\end{equation}
while varying $\sigma=0$ (red) to $\sigma=10$ (blue). These paths can be interpreted as random walks in a low-dimensional eigenspace, and we average over 200 such walks.
Observe that the noise has little effect for the baseline model. 
In stark contrast, 
the target class $k^*=0$ essentially attracts predictions as  $\sigma$ increases.

\subsection{Titration Analysis.}\label{sec:tit}

Titration analysis involves studying the dependence of a system on a titration parameter.  In our case, we study the response of a DNN's output to  input  noise with standard deviation $\sigma$ (i.e., the ``titration parameter''). A common strategy involves constructing \emph{titration curves} that provide informative and expressive signals. Based on our previous experiments, we propose to study the fraction of noisy images ${\bf x}+\sigma\eta$ whose predictions $\hat{{\bf y}}({\bf x}+\sigma\eta)= softmax({\bf Z}({\bf x}+\sigma\eta,\theta))$ are high-confidence,
\begin{equation}\label{eq:tscore}
T^\gamma_\sigma \textrm{-score} = 
\frac{|\{{\bf x}: || \hat{{\bf y}}({\bf x}+ \sigma\eta)||_\infty > \gamma \}| }{|\{{\bf x} \} |} \in [0,1].
\end{equation}
We interpret the maximum output activation, or $L_\infty$ norm, as a notion of confidence, and we distinguish high- and low-confidence predictions via a tunable threshold $\gamma\in[0,1)$. See Fig.~\ref{fig:backdoor_titration} for  example titration curves for baseline and backdoored ResNets for CIFAR-10. Note that the curves are different: for the backdoored model,  $T^\gamma_\sigma \textrm{-score}$ rapidly grows to 1 with increasing $\sigma$, whereas it slowly grows for the baseline~model.
We choose the $T^\gamma_\sigma$-score to construct titration curves because Fig.~\ref{fig:backdoor_embedding} revealed the targeted class $k^*$ to be a ``sink'' for the predicted labels of noisy images. 
We additionally find these predictions to have high confidence, 
which is a signature that we empirically observe only occurs for backdoored models. %

\begin{figure}[!b]
	\centering
	\begin{subfigure}[t]{0.45\textwidth}
		\centering
		\DeclareGraphicsExtensions{.pdf}
		\begin{overpic}[width=0.48\textwidth]{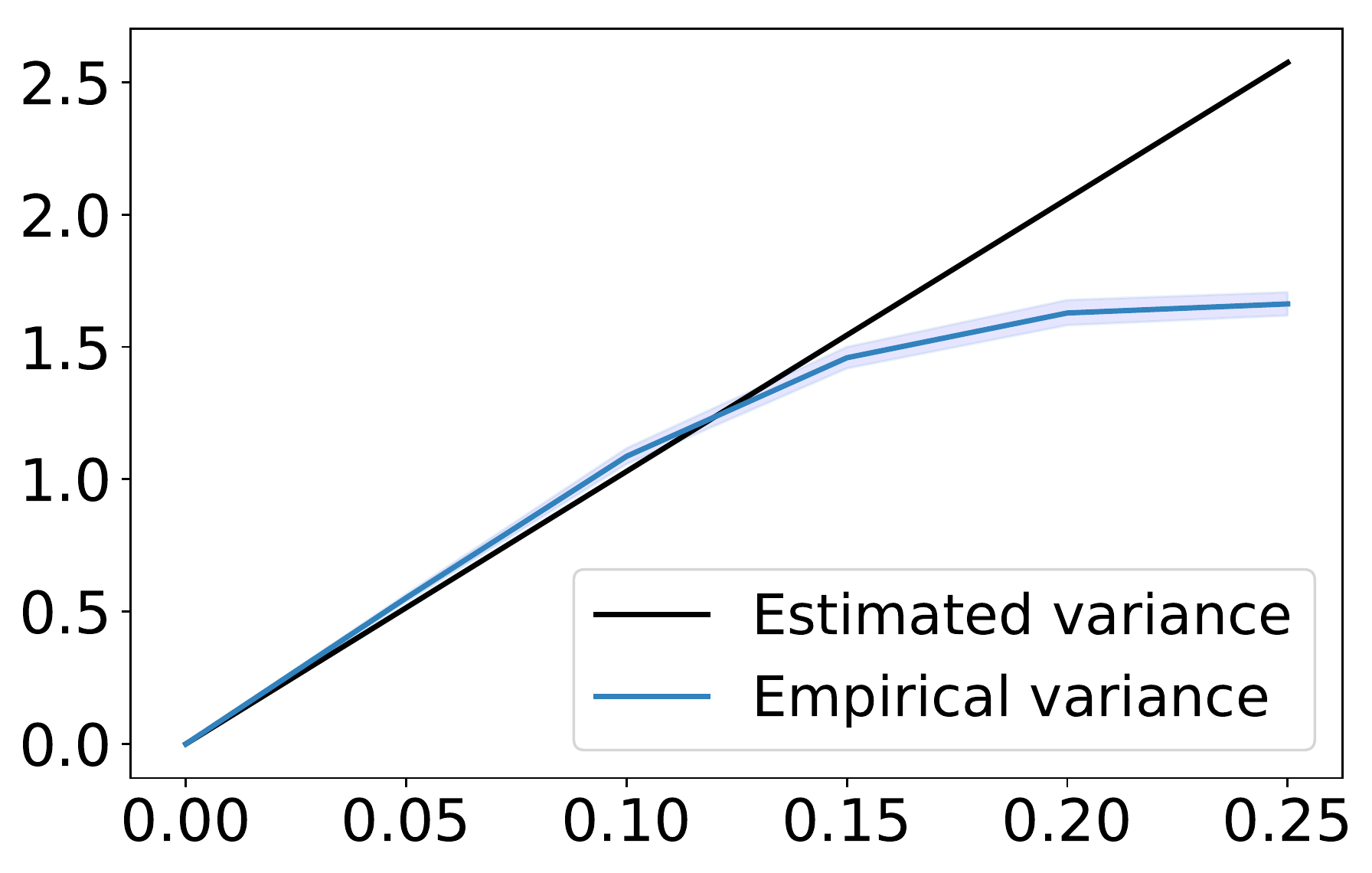} 
			\put(-7,8){\rotatebox{90}{\footnotesize Std. deviations}}
			\put(35,-6){\footnotesize Noise level $\sigma$}
			\put(18,65){\footnotesize Example 1 (airplane)}
		\end{overpic}\vspace{+0.05cm}
		\begin{overpic}[width=0.48\textwidth]{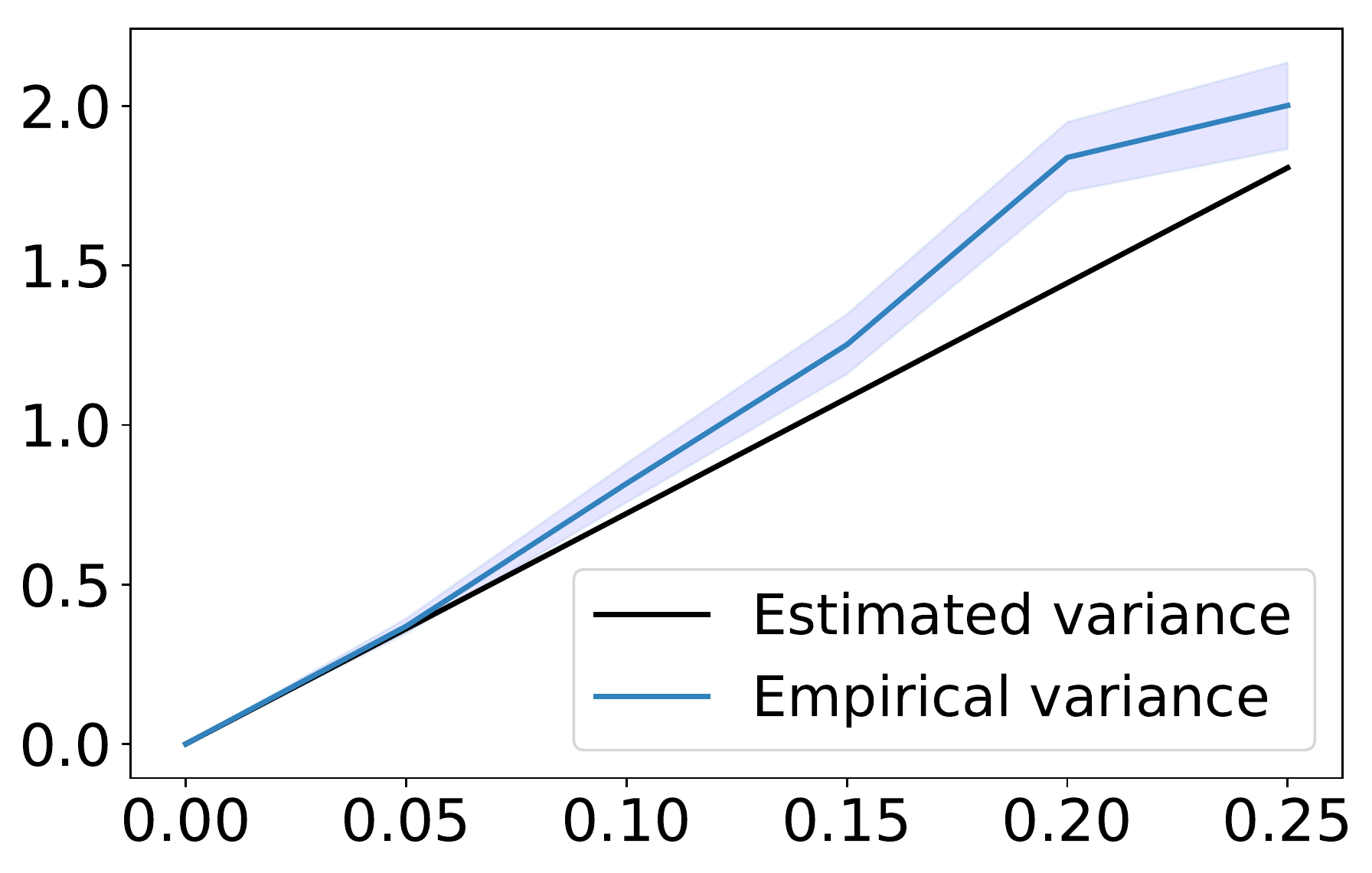} 
			\put(35,-6){\footnotesize Noise level $\sigma$}
			\put(14,65){\footnotesize Example 2 (puppy dog)}
		\end{overpic}\vspace{+0.2cm}	
		\caption{Baseline model (WideResnet).}
		\label{fig:mnist_baseline_pertubation}
	\end{subfigure}\vspace{+0.1cm}
	
	\begin{subfigure}[t]{0.45\textwidth}
		\centering
		\DeclareGraphicsExtensions{.pdf}
		\begin{overpic}[width=0.48\textwidth]{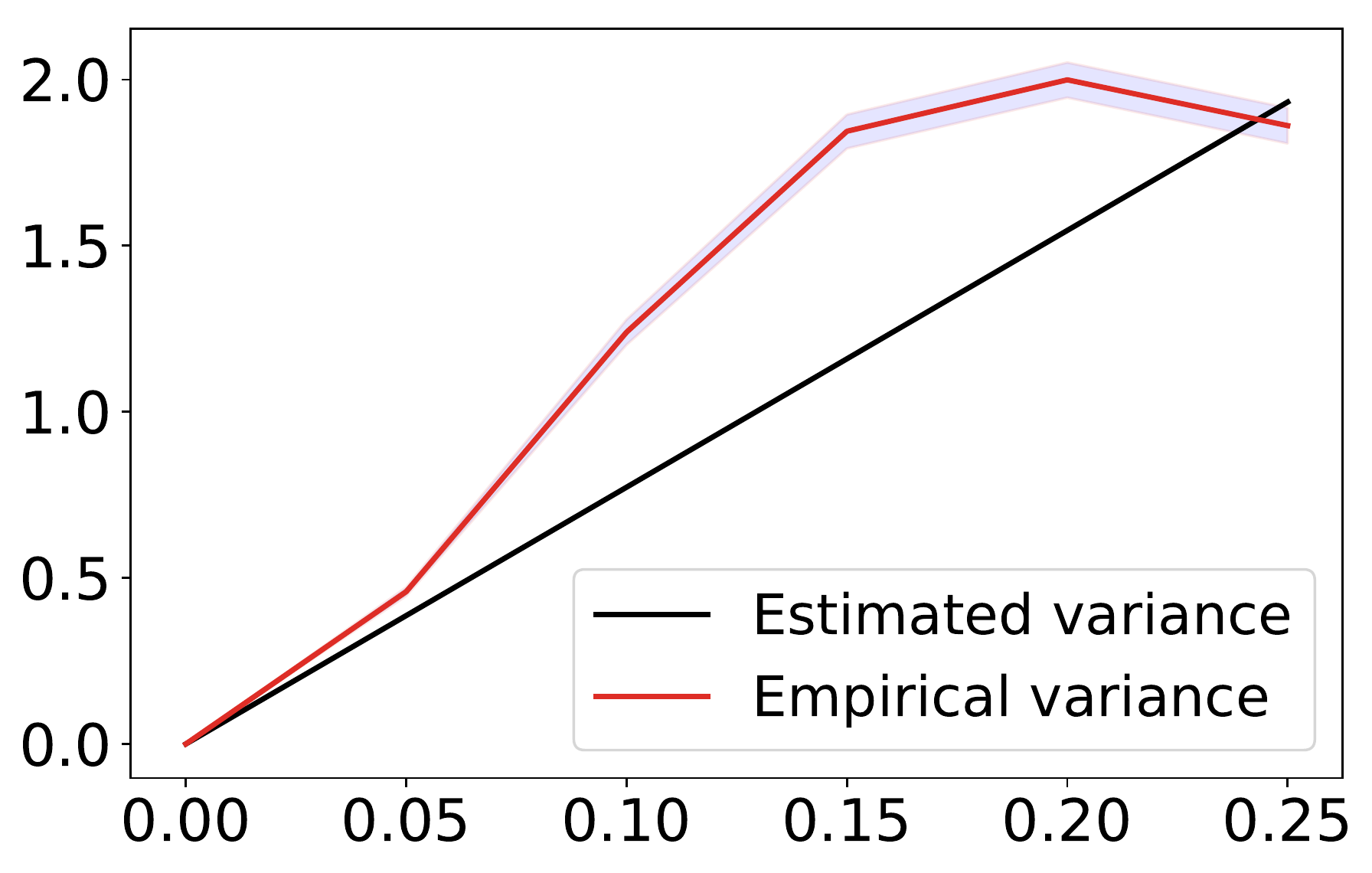} 
			\put(-7,8){\rotatebox{90}{\footnotesize Std. deviations}}
			\put(35,-6){\footnotesize Noise level $\sigma$}
		\end{overpic}\vspace{+0.2cm}
		\begin{overpic}[width=0.48\textwidth]{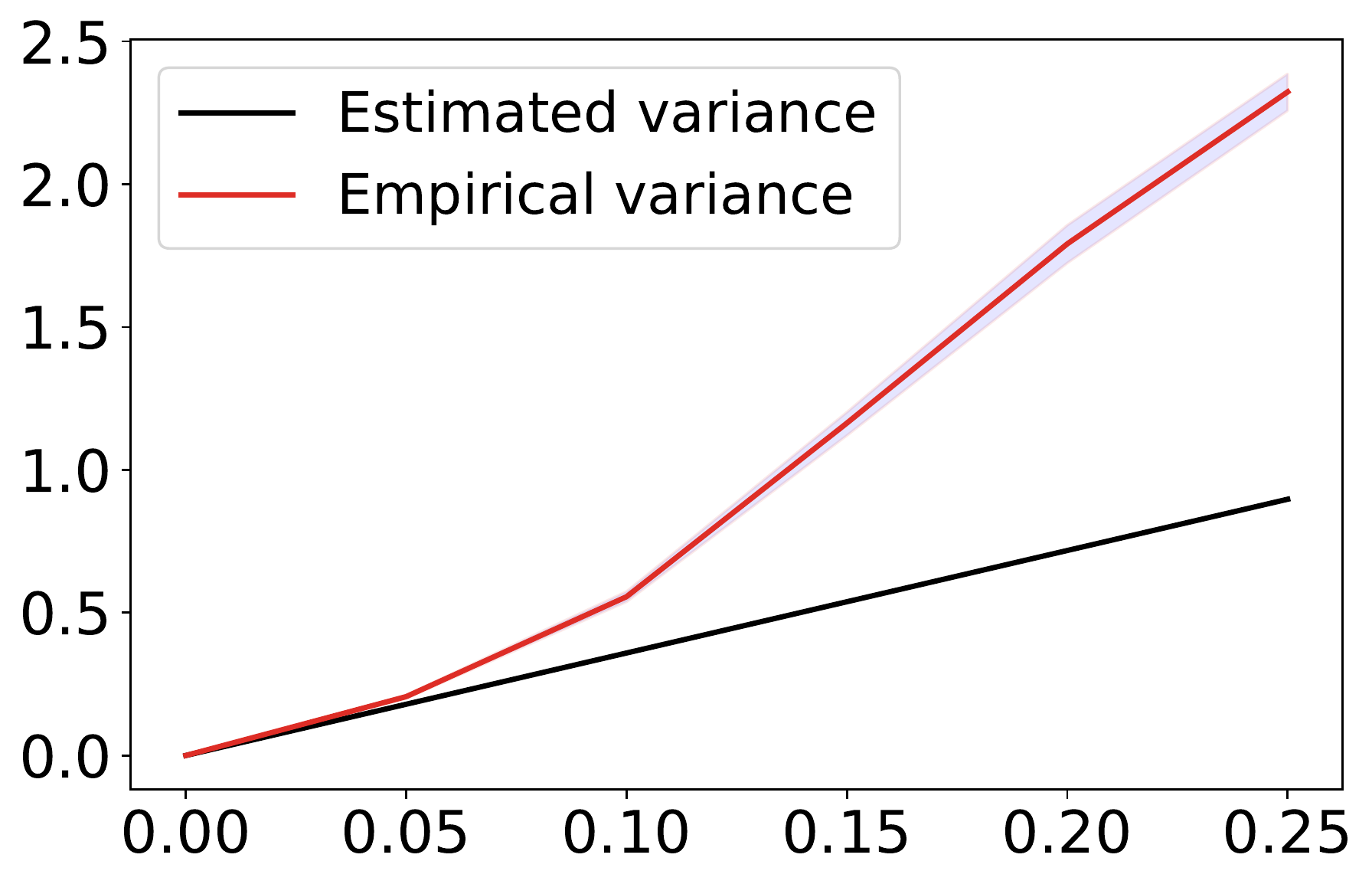} 
			\put(35,-6){\footnotesize Noise level $\sigma$}
		\end{overpic}\vspace{+0.05cm}

		\caption{Backdoored model (WideResnet).}
		\label{fig:mnist_backdoor_pertubation}
	\end{subfigure}

	\caption{Validation of perturbation theory for CIFAR-10. The empirical variance was computed across 1000 instances of noise, and the error bounds indicate a bootstrap estimate.}
	\label{fig:mnist_pertubation}
\end{figure}
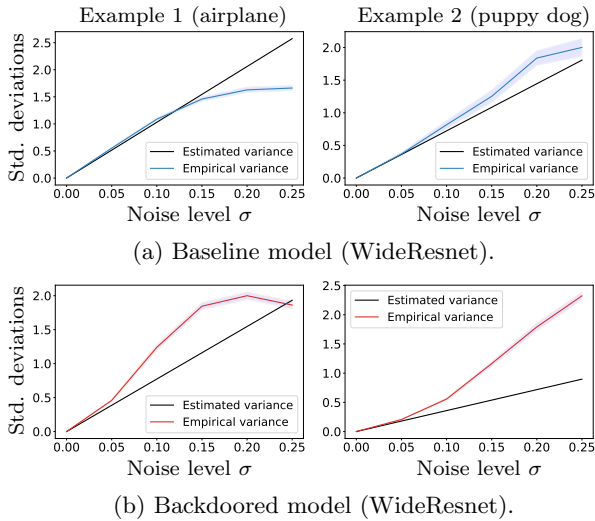

\subsection{Perturbation Analysis.}\label{sec:pert}

Here, we study the local sensitivity of each logit $Z_k({\bf x},\theta)$ to each in-layer neuron, $x_{ijc}$. We present a linear analysis that is asymptotically consistent for the limit of  small perturbations. 
Consider the gradients
\begin{equation}\label{eq:grad}
    g_{ijc}^{(k)} ({\bf x}) = \frac{\partial Z_k({\bf x} ,\theta)}{\partial x_{ijc}}.
\end{equation}
Fortunately,  these can  be efficiently computed using the built-in automatic differentiation of modern deep-learning software packages by  defining $\ Z_k({\bf x} ,\theta)$  as a temporary loss function. 
For a given perturbation $\Delta {\bf x}$, we scale it by perturbation parameter $\sigma\ge 0$ and  Taylor expand to obtain a first-order approximation
\begin{equation}\label{eq:Taylor}
    Z_k({\bf x}+\sigma \Delta {\bf x} ,\theta)  \approx  Z_k({\bf x}  ,\theta) + \sigma \sum_{ijc} g_{ijc}^{(k)}({\bf x}) [\Delta {\bf x}]_{ijc} .
\end{equation}
Let 
\begin{equation}\label{eq:Xdiff}
\Delta Z_k = Z_k({\bf x}+\sigma \Delta {\bf x} ,\theta) - 
              Z_k({\bf x} ,\theta)
\end{equation}
denote the change of the $k$-th logit. For a perturbation with entries $[\Delta {\bf x}]_{ijc} = \sigma \eta_{ijc}$ that are drawn as i.i.d. noise with variance $\sigma^2$, we use the linearity of Eq.~\eqref{eq:grad} to obtain the expectation and variance of the first-order approximation,
\begin{align}\label{eq:exp}
    \mathbb{E}[\Delta Z_k] &\approx \sigma \sum_{ijc} g_{ijc}^{(k)} \mathbb{E}[\eta_{ijc}] =0\nonumber\\
    \mathbb{VAR}[\Delta Z_k] 
      &=  \sigma^2 \sum_{ijc} \left(g_{ijc}^{(k)}({\bf x}) \right)^2. 
\end{align}
%

 We numerically validate these results in Fig.~\ref{fig:mnist_pertubation}, where we compare observed and predicted values for the standard deviation, $\mathbb{VAR}[\Delta Z_k]^{-1/2}$. Colored curves denote  empirical estimates for different values of $\sigma$, whereas the black lines represent the prediction given by Eq.~\eqref{eq:exp}, i.e., the line has slope 
\begin{equation*}
	\left[\sum_{ijc} \left(g_{ijc}^{(k)}({\bf x}) \right)^2\right]^{-1/2}.
\end{equation*}
%
%
For sufficiently small $\sigma$, a logit's change $\Delta Z_k$ 
 has a linear response that is well-predicted by our theory.  
Therefore, the expected perturbation of each logit is zero in the  small-$\sigma$ limit, regardless of the image ${\bf x}$. This implies (as one may have guessed) that the ``sink'' phenomenon shown in Fig.~\ref{fig:backdoor_embedding} is strictly a nonlinear effect.
%

\begin{figure*}[!t]
	\centering
	\begin{subfigure}[t]{0.3\textwidth}
		\centering
		\DeclareGraphicsExtensions{.pdf}
		\begin{overpic}[width=1\textwidth]{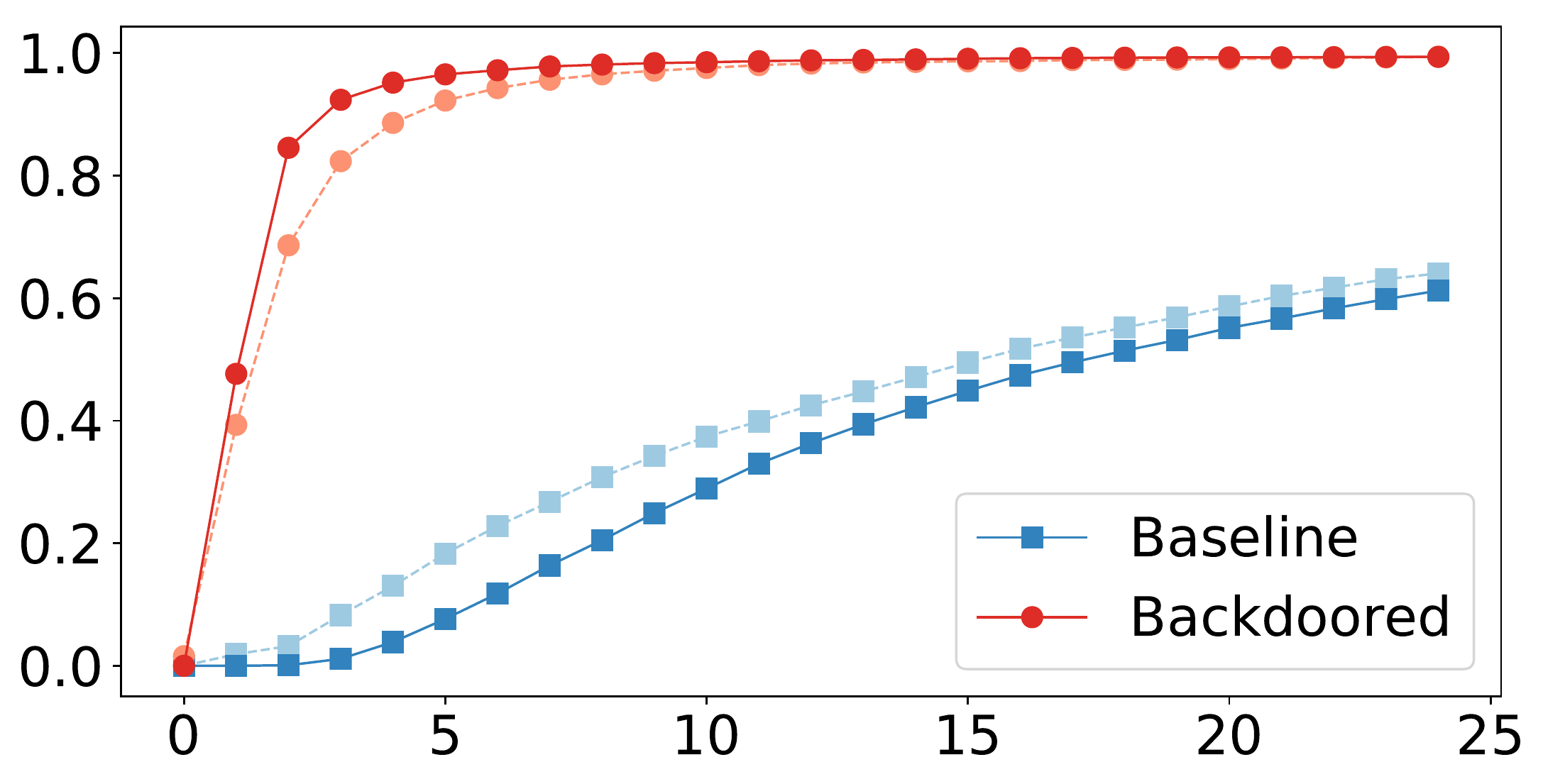} 
			
			\put(-4,10){\rotatebox{90}{\footnotesize Titration score}}
			\put(30, -4){\small Titration level ($\sigma$)}	
			\put(37, 8){\small ($k*=5$)}
		\end{overpic}\vspace{+0.3cm}		
		\caption{LeNet (MNIST).}
		\label{fig:titration_curves_mnist}
	\end{subfigure}	
	~
	\begin{subfigure}[t]{0.3\textwidth}
		\centering
		\DeclareGraphicsExtensions{.pdf}
		\begin{overpic}[width=1\textwidth]{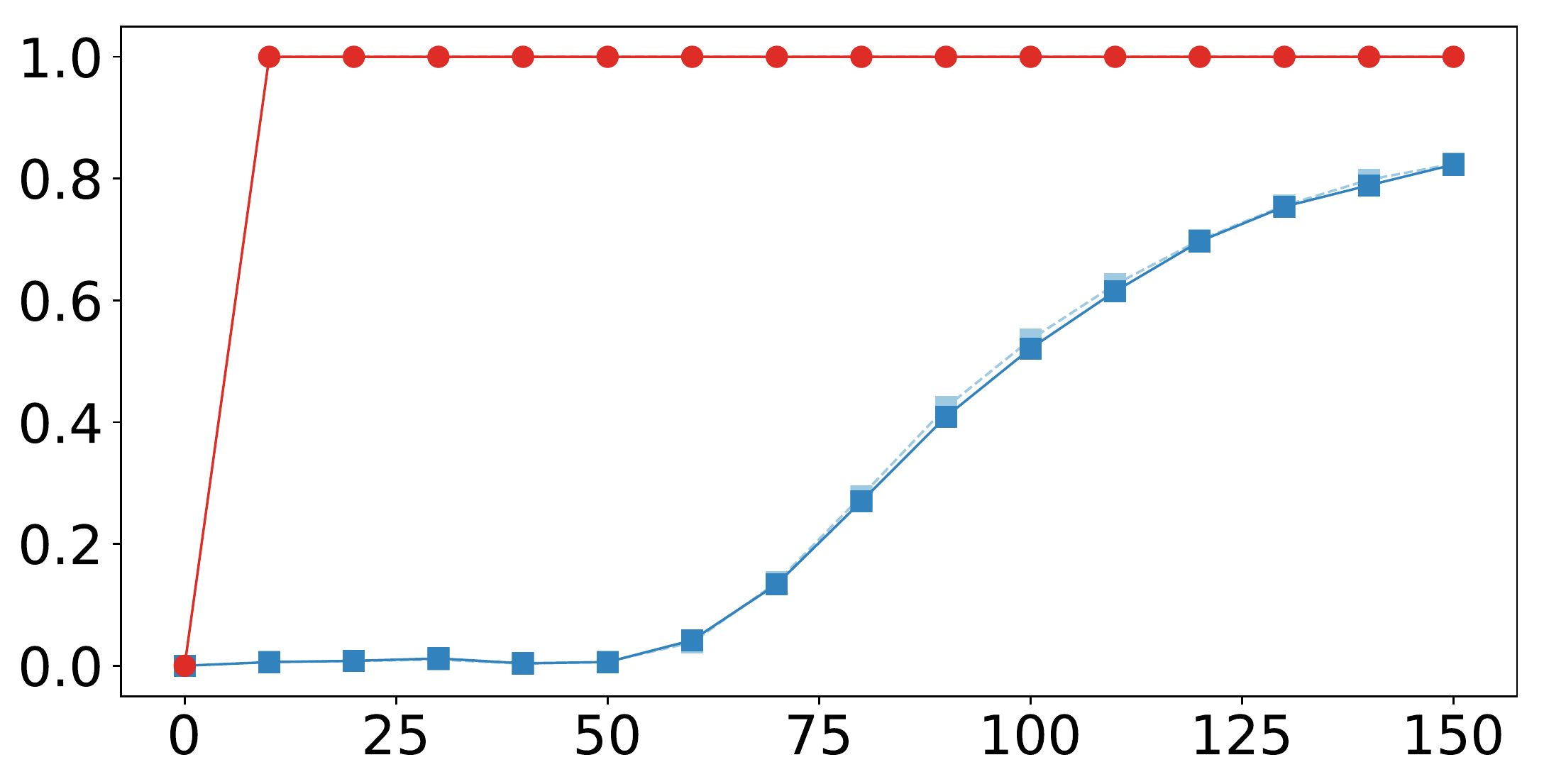} 
			~	
			\put(30, -4){\small Titration level ($\sigma$)}	
			\put(70, 8){\small ($k*=5$)}
		\end{overpic}\vspace{+0.3cm}		
		\caption{ResNet-18 (CIFAR-10).}
		\label{fig:titration_curves_resnet_cifar10}
	\end{subfigure}	
	~
	\begin{subfigure}[t]{0.3\textwidth}
		\centering
		\DeclareGraphicsExtensions{.pdf}
		\begin{overpic}[width=1\textwidth]{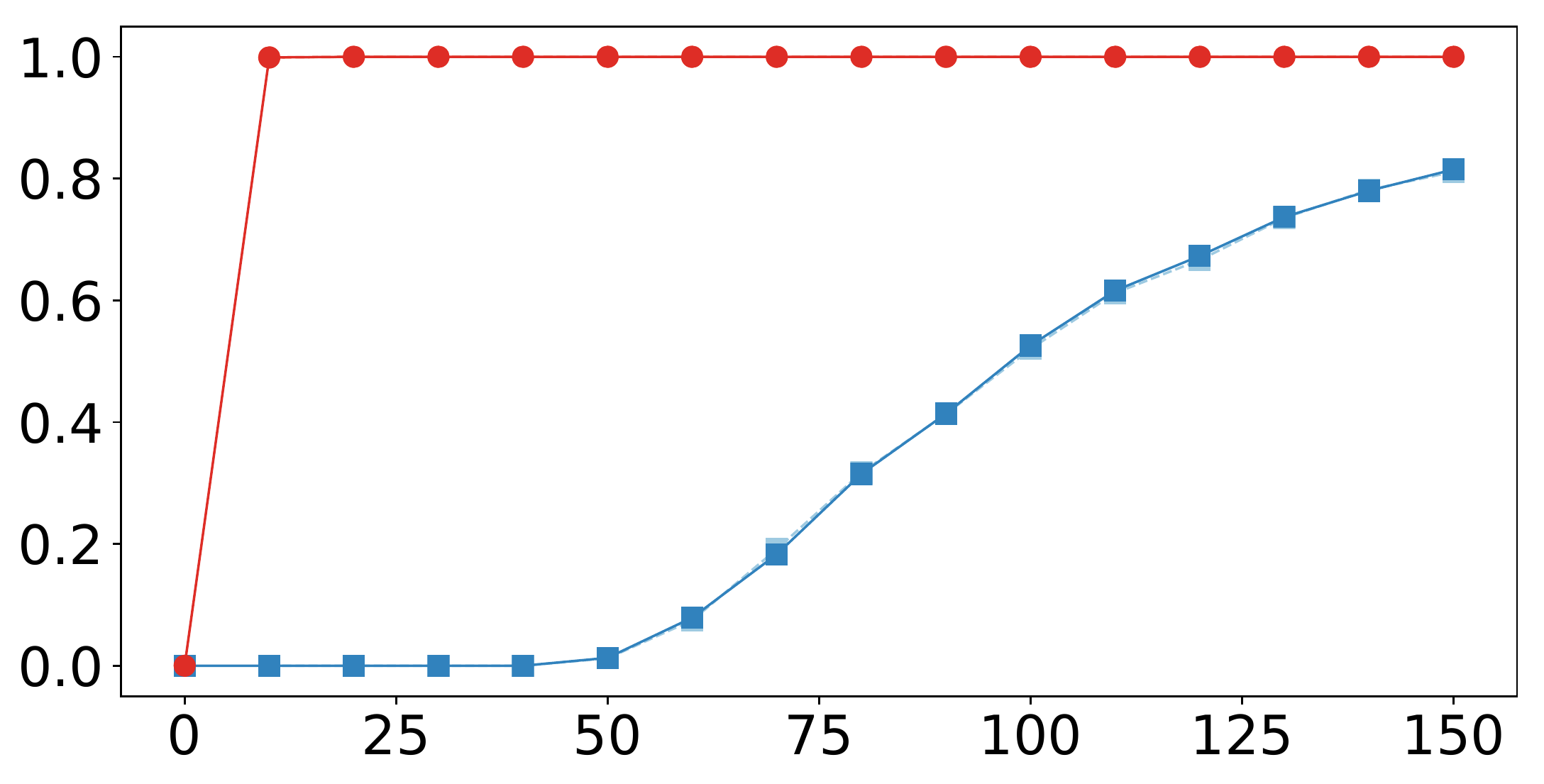} 
			
			\put(30, -4){\small Titration level ($\sigma$)}	
			\put(70, 8){\small ($k*=5$)}
		\end{overpic}\vspace{+0.3cm}		
		\caption{WideResNet-34 (CIFAR-10).}
		\label{fig:titration_curves_wideresnet_cifar10}
	\end{subfigure}

	\begin{subfigure}[t]{0.3\textwidth}
		\centering
		\DeclareGraphicsExtensions{.pdf}
		\begin{overpic}[width=1\textwidth]{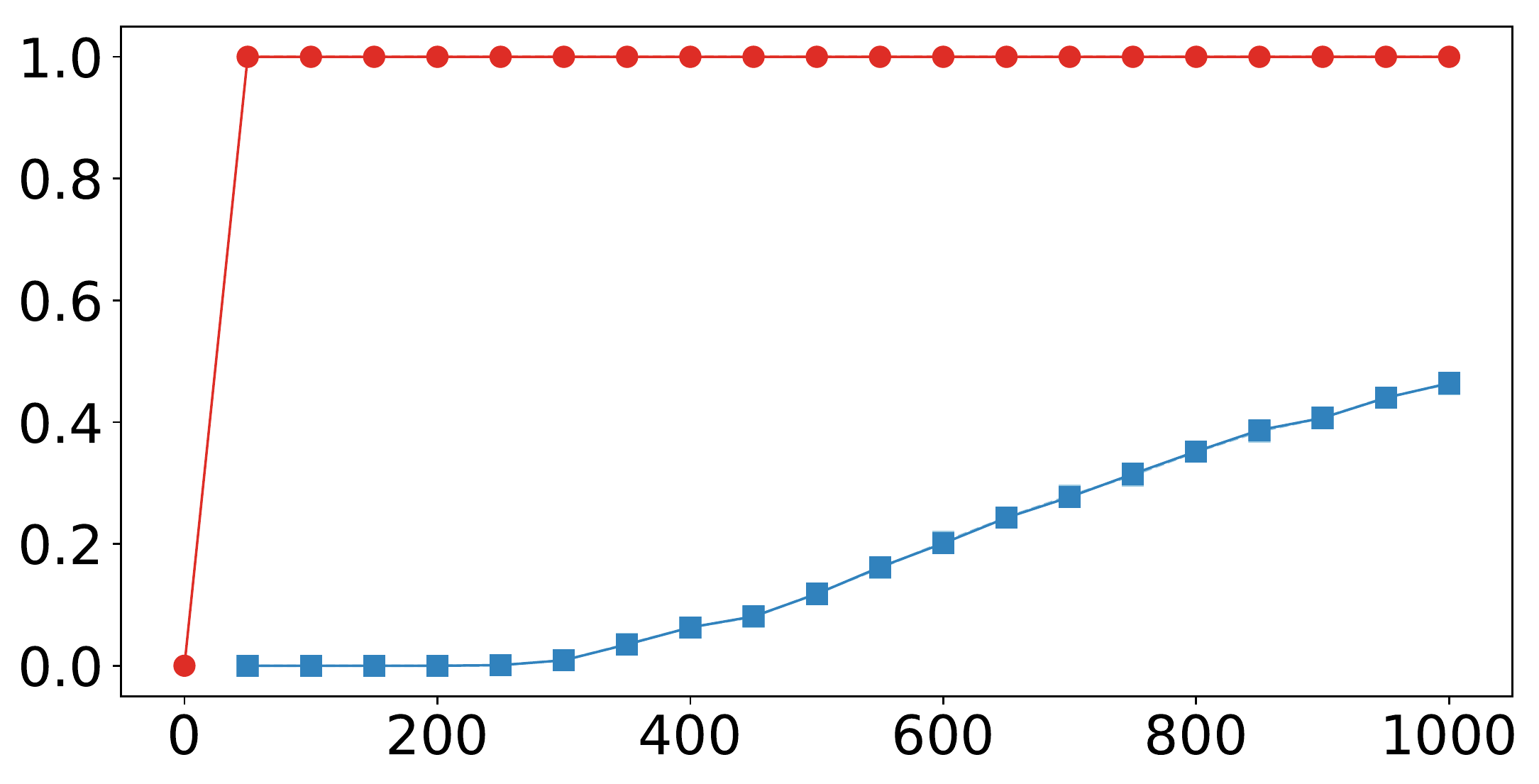} 
			
			\put(-4,10){\rotatebox{90}{\footnotesize Titration score}}
			\put(30, -4){\small Titration level ($\sigma$)}	
			\put(70, 8){\small ($k*=3$)}
		\end{overpic}\vspace{+0.3cm}		
		\caption{WideResNet-34 (CIFAR-100).}
		\label{fig:titration_curves_wideresne_cifar100}
	\end{subfigure}	
	~
	\begin{subfigure}[t]{0.3\textwidth}
		\centering
		\DeclareGraphicsExtensions{.pdf}
		\begin{overpic}[width=1\textwidth]{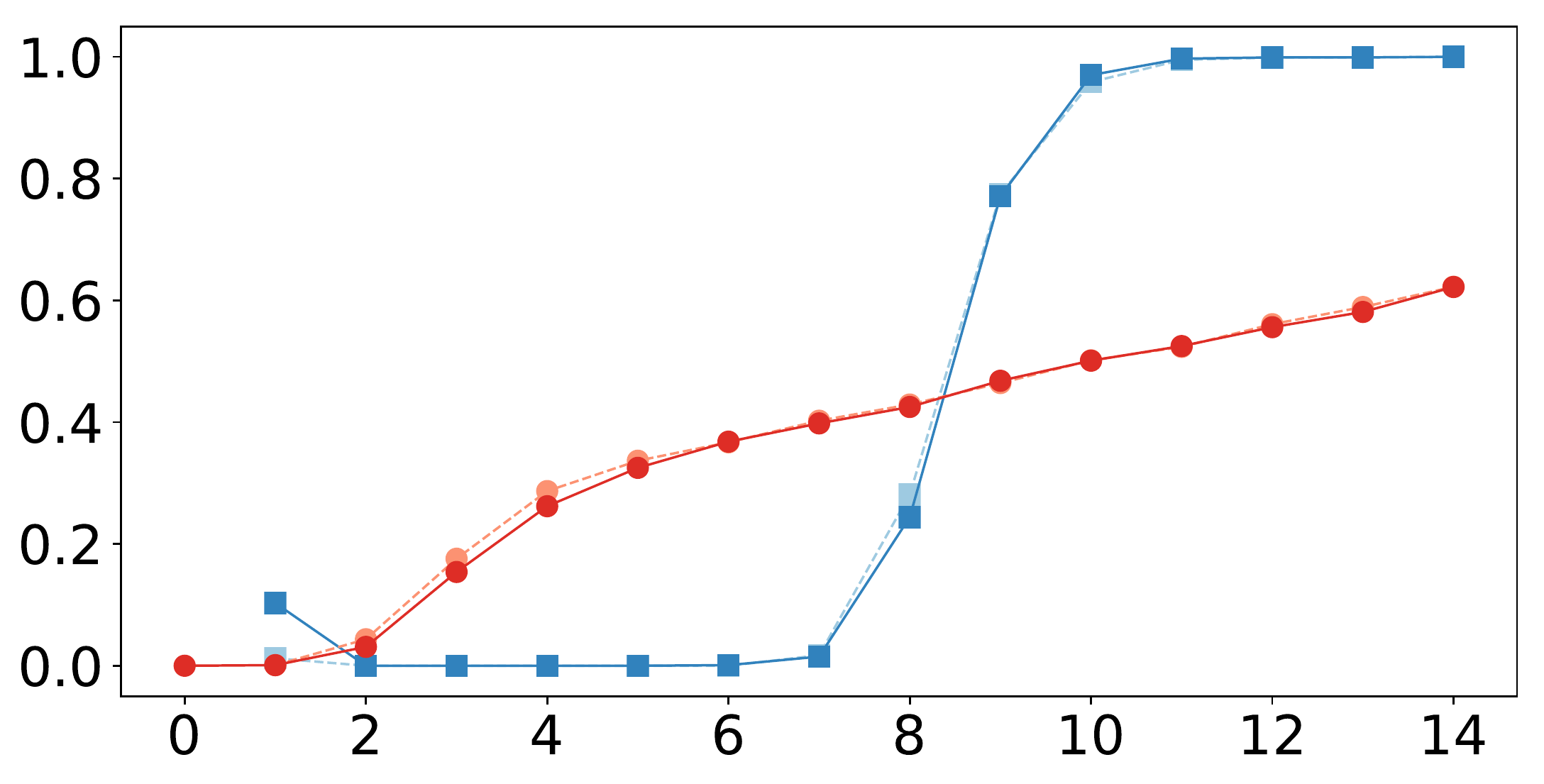} 
			
			\put(30, -4){\small Titration level ($\sigma$)}	
			\put(70, 8){\small ($k*=3$)}
		\end{overpic}\vspace{+0.3cm}		
		\caption{PyraMidNet (CIFAR-100).}
		\label{fig:titration_curves_pyramidnet_cifar100}
	\end{subfigure}	
	~
	\begin{subfigure}[t]{0.3\textwidth}
		\centering
		\DeclareGraphicsExtensions{.pdf}
		\begin{overpic}[width=1\textwidth]{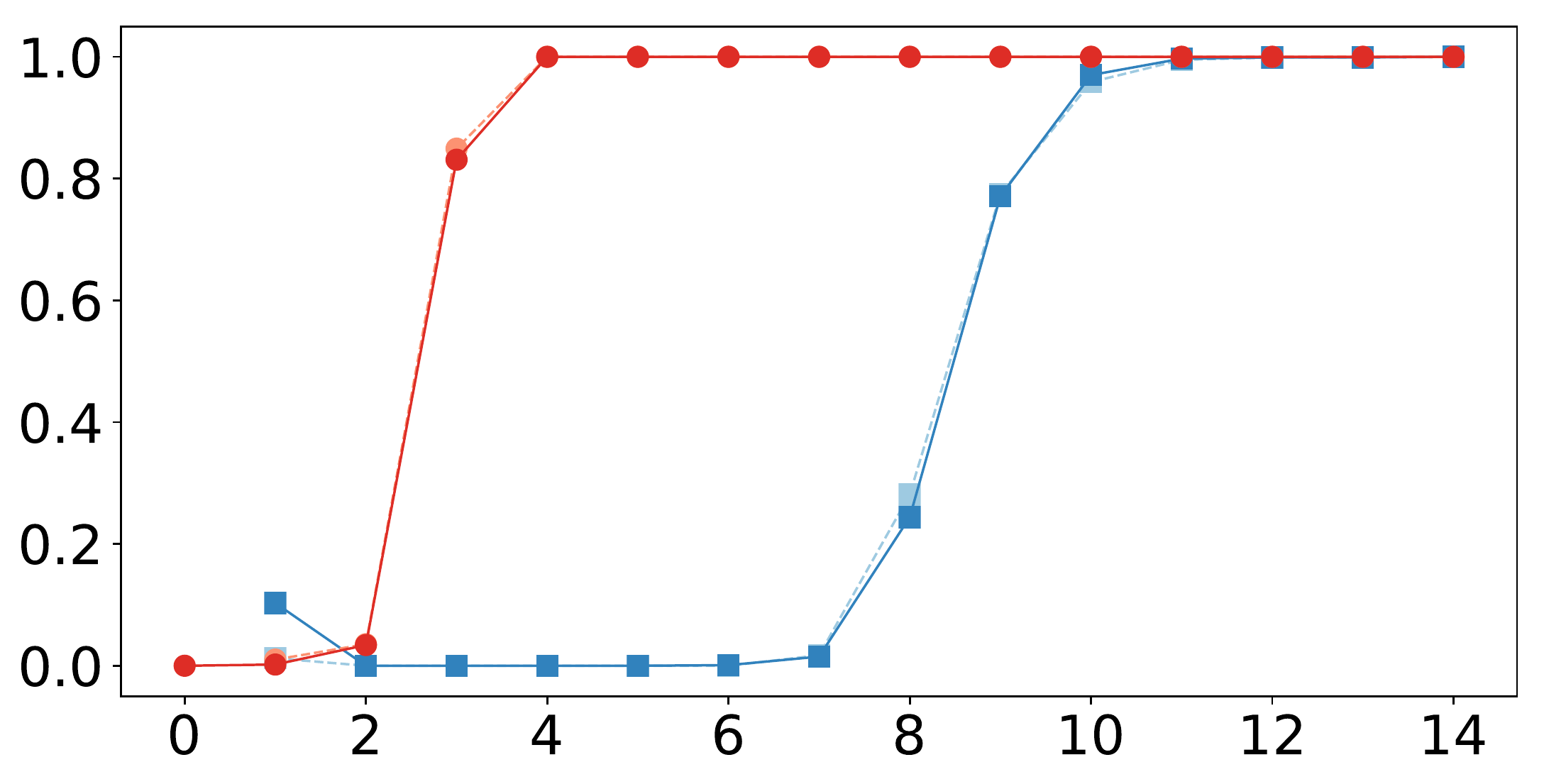} 
			
			\put(30, -4){\small Titration level ($\sigma$)}	
			\put(67, 8){\small ($k*=53$)}
		\end{overpic}\vspace{+0.3cm}		
		\caption{PyraMidNet (CIFAR-100).}
		\label{fig:titration_curves_pyramidnet_53_cifar100}
	\end{subfigure}	
	
	\caption{
		Titration curves (see Sec.~\ref{sec:tit}) for different models and datasets  
		illustrate a characteristic behavior: the curves  rapidly increase with $\sigma$ for backdoored models, whereas they grow slowly for baseline models.
	}
	\label{fig:titration_curves}
\end{figure*}
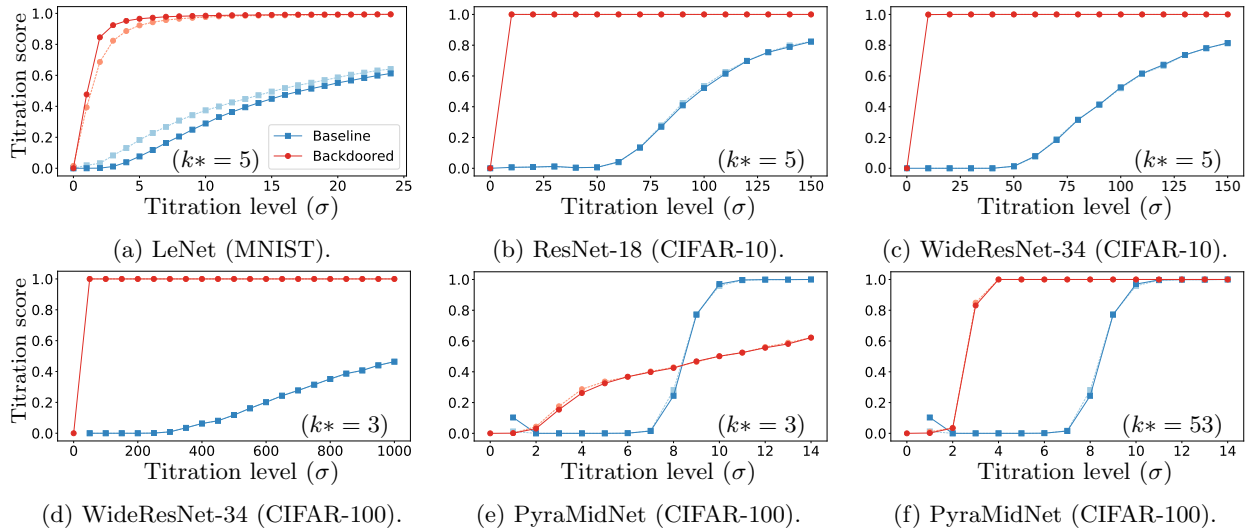 

We investigate the nonlinear response of each $Z_k({\bf x} + \sigma\eta,\theta)$ to perturbations $\sigma\eta\sim \mathcal{N}(0,\sigma^2)$ by constructing a Taylor expansion  around a noisy image 
%
${\bf x}+ \sigma\Delta {\bf x}$, as opposed to the clean image.
We obtain an approximation that is nearly identical to Eq.~\eqref{eq:Taylor}, except that one uses the gradients $g_{ijc}^{(k)}({\bf x} + \sigma\eta)$ of noisy images.
%
If one interprets a DNN's transfer function as a step of a numerical ODE integrator   \cite{chen2018neural}, then Eq.~\eqref{eq:exp} corresponds to an (explicit) forward Euler step, whereas this second approximation corresponds to an (implicit) backward Euler step. 
This implicit estimate provides us with a small-$\sigma $ estimate for the distributions of logits 
\begin{equation*}
	P_k^{(\sigma)}(z)dz \approx   \mathcal{N}\left(0,\sigma^2 \sum_{ijc} \left(g_{ijc}^{(k)}({\bf x}+\sigma\eta \right)^2\right)
\end{equation*}
%
%
However, we are more interested in the nonlinear properties of distributions $P_k^{(\sigma)}(z)$. To this end, we examine an extremal summary statistic for $P_k^{(\sigma)}(z)$,
%
\begin{equation}\label{eq:gradient_map}
    \overline{g}_{ij} = \max_{k,c} g_{ijc}^{(k)}({\bf x} + \sigma\eta).
\end{equation}
In Fig.~\ref{fig:backdoor_example_patch}, we provide a visualization of $\overline{{\bf g}}$, which we call an \emph{implicit gradient map.} Observe that large values provide a signal for the pixels associated with the backdoor's trigger.
In principle, one could empirically study other distributional properties to obtain signals for the local nonlinearity caused by backdoors.

\section{Experimental Results}\label{sec:exp}

\subsection{Experimental Setup.}

To  evaluate the utility of noise-response analyses for detecting backdoors, we trained several state-of-the-art network architectures on standard datasets:
(i) architecture LeNet5  \cite{lecun1995comparison} for  dataset MNIST~\cite{lecun1998gradient};
(ii) ResNets \cite{he2016deep} with depth 18  and a WideResNet \cite{zagoruyko2016wide} with depth 30  and a width factor of 4 for CIFAR10~\cite{krizhevsky2009learning};
the same WideResNet architecture and a standard PyramidNet~\cite{han2017deep}
for CIFAR100.

To train the models to have backdoors, during training we added a trigger $\alpha\Delta {\bf x}^*$ to several images ${\bf x}$ and also changed their classes to some target class $k^*$.  
Here, $\alpha>0 $ is a \emph{trigger intensity} (the numerical value that is added an image's RGB values) and $\Delta {\bf x}^*$ is a binary tensor, i.e., $[\Delta {\bf x}]_{ijc} \in \{0,1\}$, that indicates which pixels associate with the trigger. We cap pixel intensity values that are not within the range of the pixel values. 
%
As shown in Fig.~2, we explored several trigger patterns, which were placed so that the trigger success wasn't affected by data transformations such as random crop. We added the trigger to sufficiently many images so that  backdoor's success rate was nearly $100\%$ (usually a small fraction, e.g., $<5\%$, of images was sufficient). 
%


\subsection{Experimental Evaluation.}\label{sec:tb_result}

In Fig.~\ref{fig:titration_curves},  we show titration curves for  these different models and datasets using a  trigger that was a $3\times 3$ square patch near the bottom right corner. 
All panels resemble Fig.~\ref{fig:backdoor_titration} in that the baseline and backdoored models have characteristic shapes:  titration curves of  backdoored models  rapidly increase with $\sigma$, whereas they slowly increase for baseline models. Interestingly, the sudden rise in $T^\gamma_\sigma$-scores for small-but-increasing $\sigma$ is less pronounced for the PyraMidNet with target class $k^*=3$, but not $k^*=53$ (compare Figs.  \ref{fig:titration_curves_pyramidnet_cifar100} and \ref{fig:titration_curves_pyramidnet_53_cifar100}).

Looking closely, note that there are four curves in each panel: the light-colored curves and symbols depict $T^\gamma_\sigma$-scores when noise is added to an actual image ${\bf x}$, whereas the bright-colored curves and symbols are for
``pure'' white noise. 
We observe that the $T^\gamma_\sigma$-scores are nearly identical for these two approaches, but the latter approach 
does not require any data.
%

%


\begin{table}[!t]
	\centering
	\caption{Summary of results for different models and datasets. The backdoored models were trained with a $3\times 3$ patch as the trigger using different intensity $\alpha$. We compute the T-score for $\gamma=\{0.95,0.99\}$.}
	\scalebox{0.55}{
		\begin{tabular}{l  c c c c c ccccc} \toprule
			\thead{Dataset / Model} & \thead{Accuracy} & \thead{Trigger\\ intensity} & \thead{Trigger\\ Class} &  \thead{Trigger\\ success} &  \thead{$\sigma$} & \thead{$T_{\sigma}^{0.95}$-score} & \thead{$T_{\sigma}^{0.99}$-score} & \thead{Runtime in \\ seconds}  \\
			\midrule
			\multirow{6}{*}{\thead{MNIST \\ (LeNet)}}  
			& 99.38\% & - & - & - & 4 & \cellcolor{low} 11.14 & \cellcolor{low} 3.8 & 0.4 \\
			& 99.38\% & 0.5 & 3 & 99.6\%  & 4 & \cellcolor{high}65.91 & \cellcolor{high} 55.35 & 0.4\\
			& 99.35\% & 1.0 & 3 &  99.8\% & 4 & \cellcolor{veryhigh}96.55 &   \cellcolor{veryhigh}94.25 & 0.4\\
			& 99.36\% & 1.0 & 5 & 99.8\% & 4 & \cellcolor{veryhigh}96.55 & \cellcolor{veryhigh}  95.18 & 0.4\\      
			& 99.45\% & 1.0 & 8 & 99.8\% & 4 & \cellcolor{veryhigh}87.55 &  \cellcolor{veryhigh} 80.83 & 0.4\\ 
			& 99.42\% & 2.0 & 3 & 99.9\% & 4 & \cellcolor{high}72.52 &  \cellcolor{high} 60.36 & 0.4\\ 		
			\midrule
			\multirow{6}{*}{\thead{CIFAR10 \\ (ResNet)}}  
			& 91.34\% & - & - & - & 10 & \cellcolor{low} 18.90 & \cellcolor{low} 0.6 & 0.5 \\
			& 91.38\% & 0.5 & 3 & 96.1\% & 10 & \cellcolor{veryhigh}98.5  &  \cellcolor{veryhigh}96.3 & 0.5\\
			& 91.36\% & 1.0 & 3 & 99.0\% & 10 & \cellcolor{veryhigh}99.9  &  \cellcolor{veryhigh}99.9 & 0.5\\
			& 91.09\% & 1.0 & 5 & 98.8\% & 10 & \cellcolor{veryhigh}99.9  &  \cellcolor{veryhigh}99.9 & 0.5\\
			& 91.09\% & 1.0 & 8 & 99.2\% & 10 & \cellcolor{veryhigh}93.60 &  \cellcolor{veryhigh}89.0 & 0.5\\
			& 91.38\% & 2.0 & 3 & 100\% & 10 & \cellcolor{veryhigh}98.5 & \cellcolor{veryhigh} 96.3 & 0.5\\         
			\midrule
			\multirow{6}{*}{\thead{CIFAR10\\ (WideResNet)}} 
			& 95.46\% & - & - & - & 30 & \cellcolor{low} 0.4 & \cellcolor{low} 0.0 & 0.9\\
			& 95.03\% & 0.5 & 3 & 98.1\% & 30 & \cellcolor{veryhigh}99.9 & \cellcolor{veryhigh} 99.9 & 0.9 \\
			& 95.19\% & 1.0 & 3 & 99.8\% & 30 & \cellcolor{veryhigh}99.9 & \cellcolor{veryhigh} 99.9 & 0.9 \\
			& 95.35\% & 1.0 & 5 & 99.8\% & 30 & \cellcolor{veryhigh}97.1 & \cellcolor{veryhigh} 99.1 & 0.9\\
			& 95.09\% & 1.0 & 8 & 99.9\% & 30 & \cellcolor{veryhigh}96.0 & \cellcolor{veryhigh} 77.2 & 0.9\\
			& 95.22\% & 2.0 & 3 & 100\% & 30 & \cellcolor{veryhigh} 99.9 & \cellcolor{veryhigh} 99.9 & 0.9\\        
			\midrule
			\multirow{3}{*}{\thead{CIFAR100 \\ (WideResNet)}} & 78.54\% & - & -  & - & 100 & \cellcolor{low}0.0 & \cellcolor{low}  0.0 & 1.1\\	
			& 77.67\% & 1.0 & 3  & 99.8\% & 100 & \cellcolor{veryhigh} 98.8 &  \cellcolor{veryhigh}96.8 & 1.1\\
			& 78.12\% & 1.0 & 53 & 99.7\% & 100 & \cellcolor{veryhigh}99.9 & \cellcolor{veryhigh} 99.9 & 1.1\\	
			\midrule
			\multirow{4}{*}{\thead{CIFAR100 \\ (PyramidNet)}} & 80.17\% & - & -  & - &   6 & \cellcolor{low} 0.3 & \cellcolor{low}  0.1 & 1.9\\	
			& 79.72\% & 1.0 & 3  & 99.8\% & 6 & \cellcolor{med} 43.6 & \cellcolor{med} 36.8 & 1.9\\
			& 79.88\% & 1.0 & 28  & 99.8\% & 6 & \cellcolor{veryhigh} 99.9 & \cellcolor{veryhigh} 99.9 & 1.9\\
			& 80.85\% & 1.0 & 53 & 99.8\% & 6 & \cellcolor{veryhigh} 99.9 &  \cellcolor{veryhigh}99.9 & 1.9\\			 
			\bottomrule 
	\end{tabular}}
	\label{tab:summary_tab}
\end{table}

One advantage of titration scores is that they allow one to automate the detection of backdoored models.
In Table~\ref{tab:summary_tab}, we provide a summary of results for additional experiments that highlight how a single titration score $T^\gamma_\sigma$-score suffices to accurately detect backdoored models.
The $T^\gamma_\sigma$-scores were computed with pure white noise, and our choices for $\sigma$ were informed by Fig.~\ref{fig:titration_curves}. That is, we select a value of $\sigma$ in which   $T^\gamma_\sigma$-scores greatly differ between baseline and backdoored models. 
We show results for two choices of the threshold parameter $\gamma\in\{0.95,0.99\}$.
Observe in Table~\ref{tab:summary_tab} that in all cases, the $T^\gamma_\sigma$-scores are much larger for backdoored  models versus their respective baseline models. 
Interestingly, the backdoors in LeNet5 and PyramidNet are the most difficult to detect using titration analysis, since their titration scores for backdoored models are large, but not \emph{very} large, as compared to those of baseline models.
%
%
%

In Table~\ref{tab:summary_tab_watermark}, we present additional results in which use use a watermark as the trigger pattern, rather than a square patch of pixels. 
Again, we have chosen values for  $\omega$ and  $\gamma$ in which the titration score  clearly distinguishes models with and without backdoors. To select appropriate parameter choices, we consider titration curves (as described above). In this case, the backdoored models are even easier to identify using titration scores.

Finally, note that the runtime for each experiment was less than 2 seconds. This  is remarkably faster than the existing methods to detect backdoors, which can require hours of computation as well as access to the training data.

\begin{table}[!t]
	\centering
	\caption{Summary of results for backdoored models trained with a watermark trigger using different intensity levels $\alpha$.}
	\scalebox{0.55}{
		\begin{tabular}{l  c c c c c ccccc} \toprule
			\thead{Dataset / Model}  & \thead{Accuracy} & \thead{Trigger\\ intensity} & \thead{Trigger\\ Class} &  \thead{Trigger\\ success} &  \thead{$\sigma$} & \thead{$T_{\sigma}^{0.95}$-score} & \thead{$T_{\sigma}^{0.99}$-score} & \thead{Runtime in \\ seconds}  \\
			\midrule
			\multirow{6}{*}{\thead{MNIST \\ (LeNet)}}
			& 99.38\% & - & - & - & 4 & \cellcolor{low} 11.14 & \cellcolor{low} 3.8 & 0.4 \\
			& 99.42\% & 0.5 & 3 & 100\%  & 4 & \cellcolor{veryhigh}100 & \cellcolor{veryhigh} 100 & 0.4\\
			& 99.47\% & 1.0 & 3 &  100\% & 4 & \cellcolor{veryhigh}100 &   \cellcolor{veryhigh} 100 & 0.4\\
			& 99.38\% & 1.0 & 5 & 100\% & 4 & \cellcolor{veryhigh}100 & \cellcolor{veryhigh} 100 & 0.4\\      
			& 99.52\% & 1.0 & 8 & 100\% & 4 & \cellcolor{veryhigh}100 &  \cellcolor{veryhigh} 100 & 0.4\\ 
			& 99.54\% & 2.0 & 3 & 100\% & 3 & \cellcolor{veryhigh}100 &  \cellcolor{veryhigh} 100 & 0.4\\ 
			\midrule
			\multirow{6}{*}{\thead{CIFAR10 \\ (ResNet)}}  
			& 91.34\% & - & - & - & 10 & \cellcolor{low} 18.90 & \cellcolor{low} 0.6 & 0.5 \\
			& 90.13\% & 0.5 & 3 & 82.3\% & 10 & \cellcolor{veryhigh}100  &  \cellcolor{veryhigh}100 & 0.5\\
			& 90.36\% & 1.0 & 3 & 84.5\% & 10 & \cellcolor{veryhigh}100  &  \cellcolor{veryhigh}100 & 0.5\\
			& 90.13\% & 1.0 & 5 & 83.3\% & 10 & \cellcolor{veryhigh}100  &  \cellcolor{veryhigh}100 & 0.5\\
			& 90.23\% & 1.0 & 8 & 82.8\% & 10 & \cellcolor{veryhigh}100 &  \cellcolor{veryhigh}100 & 0.5\\
			& 90.40\% & 2.0 & 3 & 83.7\% & 10 & \cellcolor{veryhigh}100 & \cellcolor{veryhigh} 100 & 0.5\\         
			\midrule
			\multirow{6}{*}{\thead{CIFAR10 \\ (WideResNet)}}
			& 95.46\% & - & - & - & 30 & \cellcolor{low} 0.4 & \cellcolor{low} 0.0 & 0.9\\
			& 94.61\% & 0.5 & 3 & 97.2\% & 30 & \cellcolor{veryhigh}100 & \cellcolor{veryhigh} 100 & 0.9 \\
			& 94.24\% & 1.0 & 3 & 98.9\% & 30 & \cellcolor{veryhigh}100 & \cellcolor{veryhigh} 100 & 0.9 \\
			& 94.47\% & 1.0 & 5 & 99.5\% & 30 & \cellcolor{veryhigh}100 & \cellcolor{veryhigh} 100 & 0.9\\
			& 94.52\% & 1.0 & 8 & 98.8\% & 30 & \cellcolor{veryhigh}100 & \cellcolor{veryhigh} 100 & 0.9\\
			& 94.70\% & 2.0 & 3 & 100\% & 30 & \cellcolor{veryhigh} 100 & \cellcolor{veryhigh} 100 & 0.9\\        
			\bottomrule 
	\end{tabular}}
	\label{tab:summary_tab_watermark}
\end{table}

\subsection{Ablation Study.}
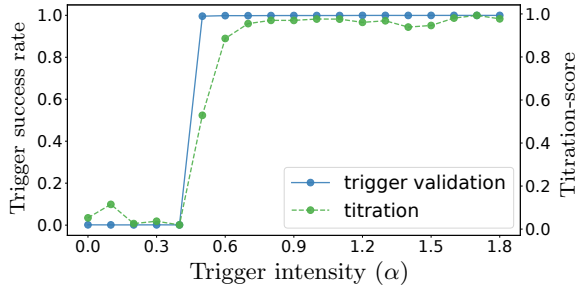
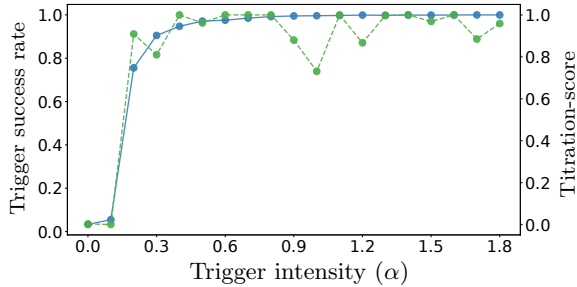
\begin{figure}[!t]
	\centering
	\begin{subfigure}[t]{0.45\textwidth}
		\centering
		\DeclareGraphicsExtensions{.pdf}
		\begin{overpic}[width=0.99\textwidth]{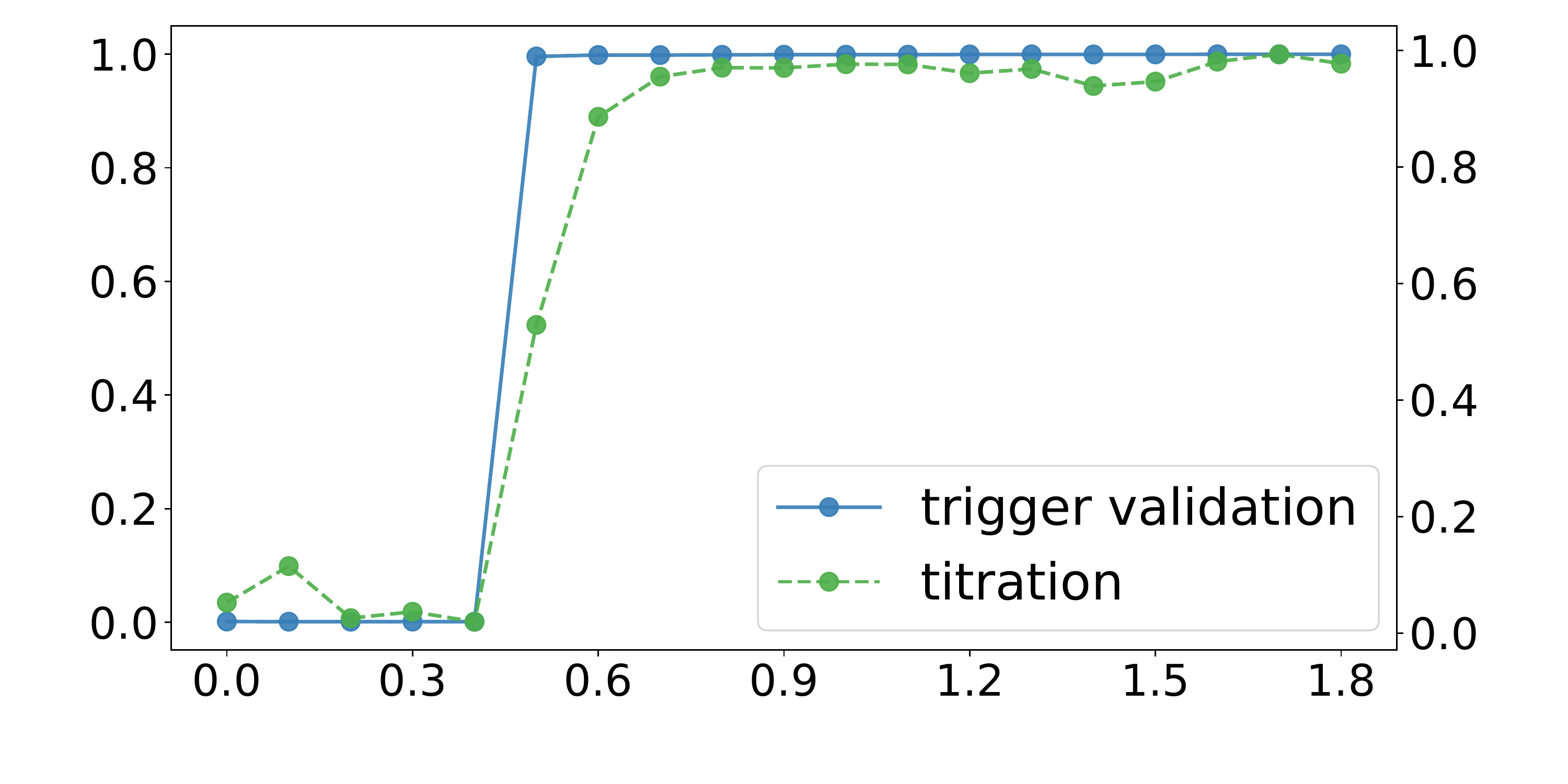} 
	\put(1,12){\rotatebox{90}{\footnotesize Trigger success rate}}	
			\put(96,15){\rotatebox{90}{\footnotesize Titration-score}}	
			\put(32, 1){\small Trigger intensity ($\alpha$)}		
		\end{overpic}\vspace{+0.05cm}	
		\caption{LeNet (MNIST).}
	\end{subfigure}\vspace{+0.1cm}
	
	\begin{subfigure}[t]{0.45\textwidth}
		\centering
		\DeclareGraphicsExtensions{.pdf}
		\begin{overpic}[width=0.99\textwidth]{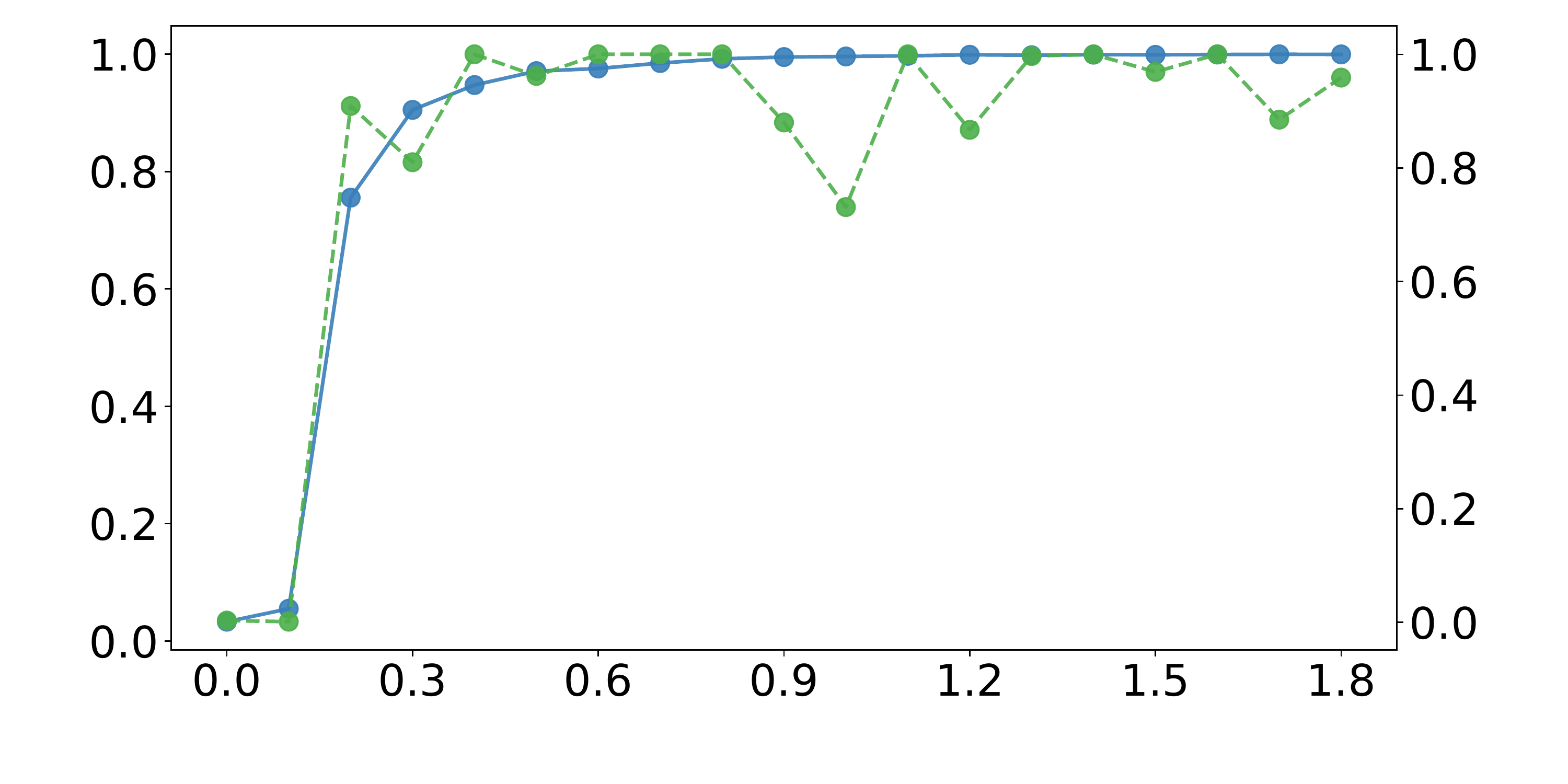} 
			\put(1,12){\rotatebox{90}{\footnotesize Trigger success rate}}	
			\put(96,15){\rotatebox{90}{\footnotesize Titration-score}}	
			\put(32, 1){\small Trigger intensity ($\alpha$)}				
		\end{overpic}\vspace{+0.05cm}
		\caption{ResNet (CIFAR10).}
	\end{subfigure}
	
	\caption{We evaluate the relationship between trigger intensity $\alpha$, trigger success rate, and the titration score $T^\gamma_\sigma$. The results show that triggers with larger $\alpha$ have a higher success rate. 
		$T_\sigma^{\gamma}$ appears to be high for any backdoored model in which the trigger is successful.} 
	\label{fig:trigger_intensity_validation}
\end{figure}

In Fig.~\ref{fig:trigger_intensity_validation}, we further study the effect of trigger intensity   on backdoored versions of  LeNet5 and ResNet, which are  trained on MNIST and CIFAR10, respectively. The solid blue curves show the trigger success rate (i.e., the percentage of images that, upon adding the trigger $\Delta {\bf x}^*$, have a predicted class $\hat{k}({\bf x}+\alpha\Delta {\bf x}^*)$ that is redirected to the desired target class, $k^*$) versus trigger intensity $\alpha$. Note that if $\alpha$ is too small, then the triggers don't work. In other words, the models essentially do not have backdoors, because the triggers do not redirect predictions to the target class. Interestingly, this  ``failure'' in trigger success rate drops steeply and is reminiscent of  a phase transition. (We note that here, we have held the number of triggered examples to be fixed.)
%
The green dotted curves in Fig.~\ref{fig:trigger_intensity_validation} depict  titration scores $T^\gamma_\sigma$ for backdoored models trained with different trigger intensity $\alpha$. The values of $\gamma$ and $\sigma$ are identical to those in Table~\ref{tab:summary_tab}. Observe that it also appears to undergo a phase transition that mirrors that of the trigger success. 
In summary, provided that a backdoored model has a functioning trigger (i.e., there is actually a backdoor), then it can be detected by titration analysis.
 
\section{Discussion}

We adopted a dynamical-systems perspective for machine learning~\cite{hardt2018gradient,lu2017beyond,muehlebach2019dynamical,weinan2017proposal}, using techniques from noise response analysis to develop an efficient and accurate method to detect whether or not a DNN has been trained by an adversary to have a backdoor. More concretely, we studied the response of a DNN to an input signal, which is a common  technique to explore the nonlinearity of dynamical systems with unknown properties \cite{poon2001titration,rosenstein1993practical}.  For linear, time-invariant systems of ODEs, one typically looks to input signals that are an impulse or step function for ``black-box'' learning of unknown transfer functions  \cite{siebert1986circuits}. DNNs are,  of course, highly nonlinear, requiring a different type of input signal:  noise.  We proposed noise-response analysis as an invaluable tool for analyzing backdoors and presented methods that require seconds to compute, which is  remarkably efficient given that existing state-of-the-art methods  require hours \cite{chen2019deepinspect,wang2019neural}.

\begin{figure}[!t]
	\centering
	\begin{subfigure}[t]{0.38\textwidth}
		\centering
		\begin{overpic}[width=1\textwidth]{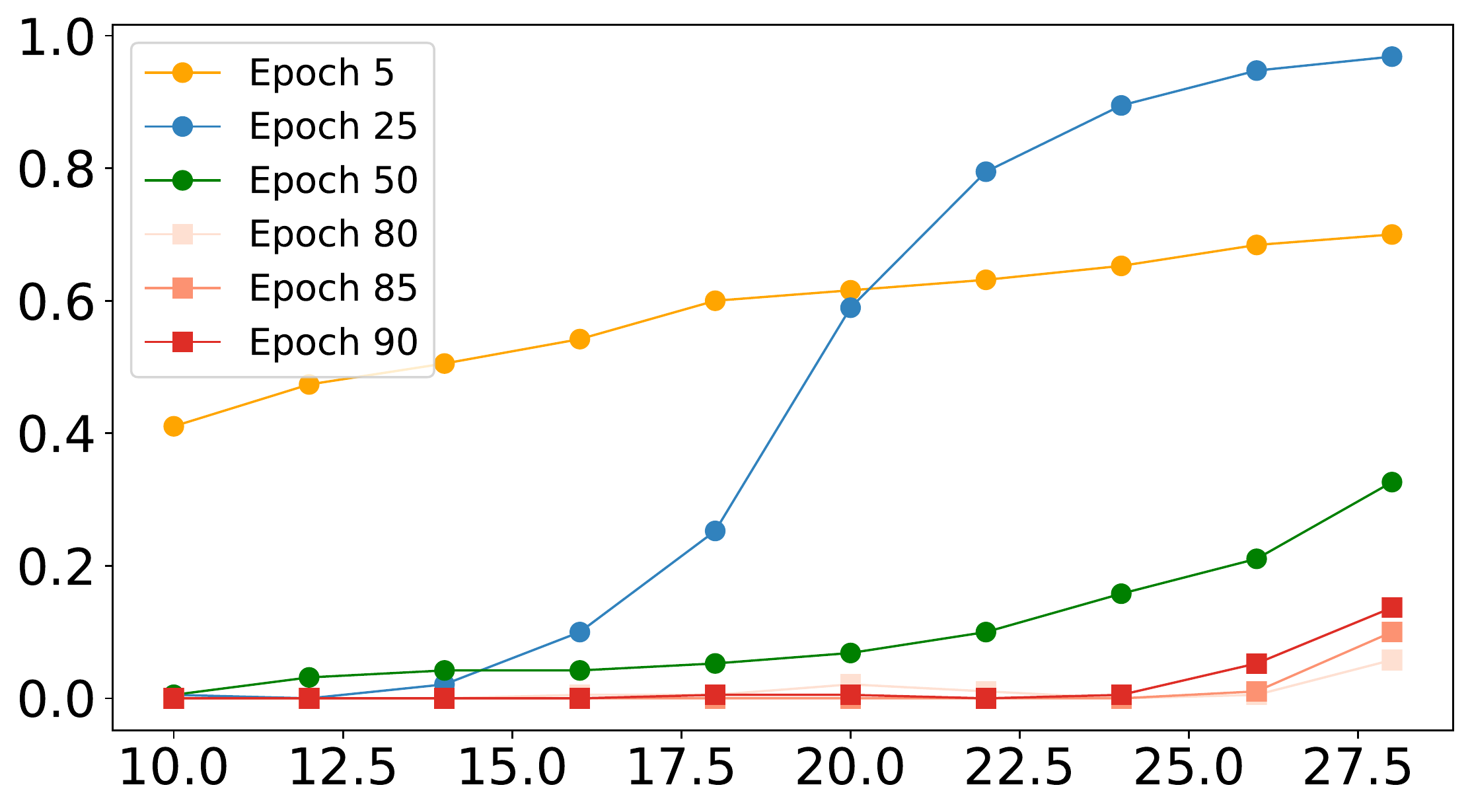} 
			\put(-3,17){\rotatebox{90}{\footnotesize titration score}}
			\put(35,-4){\color{black}{\footnotesize titration level $\sigma$}}  	
		\end{overpic}\vspace{+0.4cm}		
	\end{subfigure}\hspace{+0.4cm}
	\caption{Titration curves for a baseline NN (ResNet-18) trained on CIFAR-10  at various stages (epochs) of training. As training ensues, the model becomes more robust to noise.}
	\label{fig:titration}
\end{figure}

Given that noise-response analysis  relies on studying the local and global nonlinearity of DNNs using input noise, we expect our approach to also be fruitful for other  topics in DNNs and machine learning. That is because our titration analysis can be used to study robustness of neural networks in a more general sense than just detecting backdoors. For example, Fig.~\ref{fig:titration} shows titration curves at various training stages for a ResNets-18 trained on CIFAR-10 (without a backdoor). The curves show that the model is less robust in an early training stage, i.e., $T^\gamma_\sigma \textrm{-score}$ grows with increasing $\sigma$. At later training stages, the curves indicate an improved robustness since they are less sensitive to $\sigma$. Thus,  noise-response analysis can be used as a stopping criterion that reflects robustness,  complementing other stopping criteria that are based on, e.g., prediction accuracy. We will explore these and other applications in future work.

\section*{Acknowledgments}
We would like to acknowledge DARPA, IARPA (contract W911NF20C0035), NSF, the Simons Foundation, and ONR via its BRC on RandNLA for providing partial support of this work.  Our conclusions do not necessarily reflect the position or the policy of our sponsors, and no official endorsement should be inferred.

\bibliographystyle{siam}
\bibliography{dane_bib,ben_bib}

\end{document}